\documentclass[10pt,journal,compsoc]{IEEEtran}
\ifCLASSOPTIONcompsoc
  \usepackage[nocompress]{cite}
\else
  \usepackage{cite}
\fi

\usepackage{amsmath}
\usepackage{epsfig}
\usepackage{graphicx}
\usepackage{amssymb}
\usepackage{bm}
\usepackage{comment}
\usepackage[mathscr]{eucal}
\usepackage[dvipsnames]{xcolor}

\usepackage[normalem]{ulem}

\usepackage{enumitem}
\usepackage{booktabs}
\usepackage{array,multirow}

\newcommand{\STAB}[1]{\begin{tabular}{@{}c@{}}#1\end{tabular}}

\newcommand{\ignore}[1]{}

\newcommand{\revised}[1][black]{\textcolor{#1}}
\newcommand{\revisedminor}[1][black]{\textcolor{#1}}

\usepackage{placeins}
\usepackage[pagebackref=true,breaklinks=true,letterpaper=true,colorlinks,bookmarks=false]{hyperref}

\usepackage{xspace}
\makeatletter
\DeclareRobustCommand\onedot{\futurelet\@let@token\@onedot}
\def\@onedot{\ifx\@let@token.\else.\null\fi\xspace}

\def\eg{\emph{e.g}\onedot} 
\def\ie{\emph{i.e}\onedot} 
\def\cf{\emph{c.f}\onedot} 
\def\etc{\emph{etc}\onedot}

\makeatother

\begin{document}
%
\title{Looking Beyond Two Frames: \\ End-to-End Multi-Object Tracking Using \\ Spatial and Temporal Transformers}

\author{
Tianyu~Zhu,
Markus~Hiller,
Mahsa~Ehsanpour,
Rongkai~Ma, 
Tom~Drummond, \\
Ian~Reid
and~Hamid~Rezatofighi
\IEEEcompsocitemizethanks{
\IEEEcompsocthanksitem 
T. Zhu and R. Ma are with the Department of Electrical and Computer Systems Engineering, Monash University.
\IEEEcompsocthanksitem M. Hiller and T. Drummond are with the School of Computing and Information Systems, The University of Melbourne. 
\IEEEcompsocthanksitem M. Ehsanpour and I. Reid are with the Australian Institute for Machine Learning, The University of Adelaide. 
\IEEEcompsocthanksitem H. Rezatofighi is with the Department of Data Science and AI, Monash University. 
\IEEEcompsocthanksitem TZ, MH, ME, RM, TD and IR are associated with the Australian Centre for Robotic Vision. 
\IEEEcompsocthanksitem Corresponding authors: Markus Hiller (mhiller@student.unimelb.edu.au); Tianyu Zhu (tianyu.zhu@monash.edu)
}
}

\IEEEtitleabstractindextext{%
\begin{abstract}
Tracking a time-varying indefinite number of objects in a video sequence over time remains a challenge despite recent advances in the field. Most existing approaches are not able to properly handle multi-object tracking challenges such as occlusion, in part because they ignore long-term temporal information.
To address these shortcomings, we present MO3TR: a truly end-to-end Transformer-based online multi-object tracking (MOT) framework that learns to handle occlusions, track initiation and termination without the need for an explicit data association module or any heuristics. 
MO3TR encodes object interactions into long-term temporal embeddings using a combination of spatial and temporal Transformers, and recursively uses the information jointly with the input data to estimate the states of all tracked objects over time. The spatial attention mechanism enables our framework to learn implicit representations between all the objects and the objects to the measurements, while the temporal attention mechanism focuses on specific parts of past information, allowing our approach to resolve occlusions over multiple frames. 
Our experiments demonstrate the potential of this new approach, achieving results on par with or better than the current state-of-the-art on multiple MOT metrics for several popular multi-object tracking benchmarks.
\end{abstract}

}

\maketitle

\IEEEdisplaynontitleabstractindextext

%
\IEEEpeerreviewmaketitle
\IEEEraisesectionheading{\section{Introduction}\label{Introduction}}%
\IEEEPARstart{V}{isually} discriminating the identity of multiple objects in a scene and creating individual \textit{tracks} of their movements over time, namely \textit{multi-object tracking}, is one of the basic yet most crucial vision tasks, imperative to tackle many real-world problems in surveillance, robotics/autonomous driving, health and biology. 
While being a rather classical AI problem, it is still very challenging to design a 
reliable multi-object tracking (MOT) system capable of tracking an unknown and time-varying number of objects moving through unconstrained environments, directly from spurious and ambiguous measurements and in presence of many other complexities such as occlusion, detection failure and data (measurement-to-objects) association uncertainty.\par
Early frameworks approached the MOT problem by splitting it into multiple sub-problems that could be targeted individually, commonly starting with object detection, followed by data association, track management and an additional filtering or state prediction stage; each with their own set of challenges and solutions~\cite{andriyenko2011multi,andriyenko2012discrete,sort,blackman1999designMHT,fortmann1983_JPDAF, rezantofighi2015_jpdarev,smith2019dataassociationreview,streit1994PMHT}.  
Recently, deep learning has considerably contributed to improving the performance of multi-object tracking approaches, but surprisingly not through learning the entire problem end-to-end. Instead, the developed methods adopted the traditional problem split of early methods and mainly focused on enhancing some of the aforementioned components, such as creating better detectors~\cite{felzenszwalb2009object,redmon2017yolo9000,ren2017accurate,NIPS2015_14bfa6bb,yang2016exploit} or developing more reliable matching objectives for associating detections to existing object tracks~\cite{hu2019joint,leal2016learning,sadeghian2017tracking,wang2020motjde,deepsort,wang2021rtu,stadler2021tmoh}. While this tracking-by-detection paradigm has become the de facto standard approach for MOT, it has its own limitations. Recent approaches have shown advances by considering detection and tracking as a joint learning task rather than two separate sequential problems~\cite{ tracktor,feichtenhofer2017detect,sun2019deep,zhou2020centertrack}. 
However, these methods often formulate the MOT task as a two consecutive frames problem and ignore long-term temporal information, which is imperative for tackling key challenges such as track initiation, termination and occlusion handling.\par
In addition to their aforementioned limitations, all these methods can barely be considered to be end-to-end multi-object frameworks as their final outputs, \ie tracks, are generated through a non-learning process. For example, track initiation and termination are commonly tackled by applying different heuristics, and the track assignments are decided upon by applying additional optimization methods, \eg the Hungarian algorithm~\cite{kuhn1955hungarian}, max-flow min-cut~\cite{ford_fulkerson_1956}, \etc, and the generated tracks may be smoothed by a process such as interpolation or filtering~\cite{kalman1960new}.\par
With the recent rise in popularity of Transformers~\cite{attentionisallyouneed,khan2021transformers}, this rather new deep learning tool has been adapted to solve computer vision problems like object detection~\cite{carion2020detr} by building on top of advances in using deep architectures for set prediction tasks~\cite{rezatofighi2018_setarxiv,rezatofighi2021_settpami}. Concurrent to our work, this idea has also been deployed to some new MOT frameworks~\cite{trackformer,sun2020transtrack}.  Nonetheless, these methods still either rely on conventional heuristics, \eg Intersection over Union (IoU) matching~\cite{sun2020transtrack}, or formulate the problem as a two-frames task~\cite{trackformer,sun2020transtrack}, making them naive approaches to handle long-term occlusions. \par
\begin{figure}[t]
    \begin{center}
    {
        \includegraphics[width=1\linewidth]{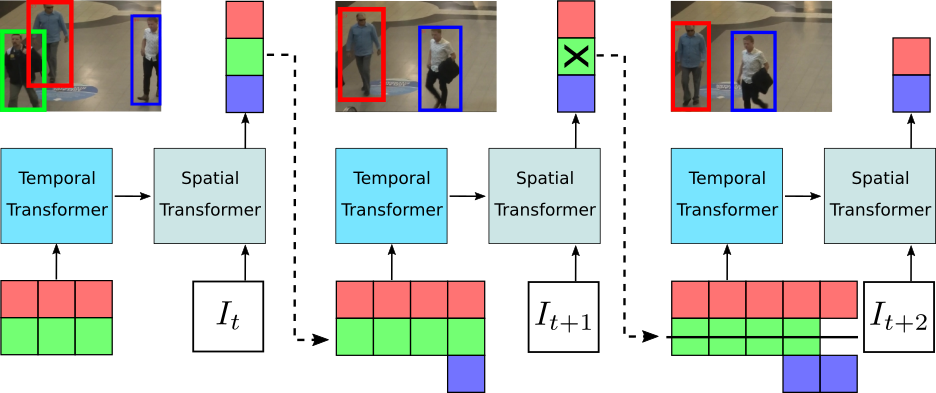}
    }
    \end{center}
    \vspace{-.5em}
  \caption{Looking beyond two frames with MO3TR: Temporal and spatial Transformers jointly pay attention to the current image~$I_t$ and the entire embedding history of the two tracked objects (\textit{red} and \textit{green}, left). Detection of a previously untracked object (\textit{blue}) causes initiation of new track (left \textrightarrow$\,$ middle), while an object exiting the scene (\textit{green}) leads to track termination (middle \textrightarrow$\,$ right). Embeddings encoding spatial and temporal interactions are accumulated over time to form individual object-track histories.}
  \vspace{-.5em}
  \label{fig:conceptual}
\end{figure}

In this paper, we show that the MOT problem can be learnt end-to-end without the use of heuristics. Our proposed method \textit{MO3TR}: \textit{\underline{M}ulti-\underline{O}bject \underline{TR}acking using spatial \underline{TR}ansformers and temporal \underline{TR}ansformers} addresses the key tasks of track initiation and termination as well as occlusion handling. Being a truly end-to-end Transformer-based online multi-object tracking method, MO3TR learns to recursively predict the state of the objects directly from an image sequence stream and encodes long-term temporal information to estimate the states of all objects over time~(Fig.~\ref{fig:conceptual}). In contrast to most existing MOT frameworks, this state-based approach does not require any explicit data association module (see Section~\ref{sec:dataassoc} for a more detailed discussion).\par
\begin{figure*}[t]
\begin{center}
\scalebox{0.9}{
  \includegraphics[width=1\linewidth]{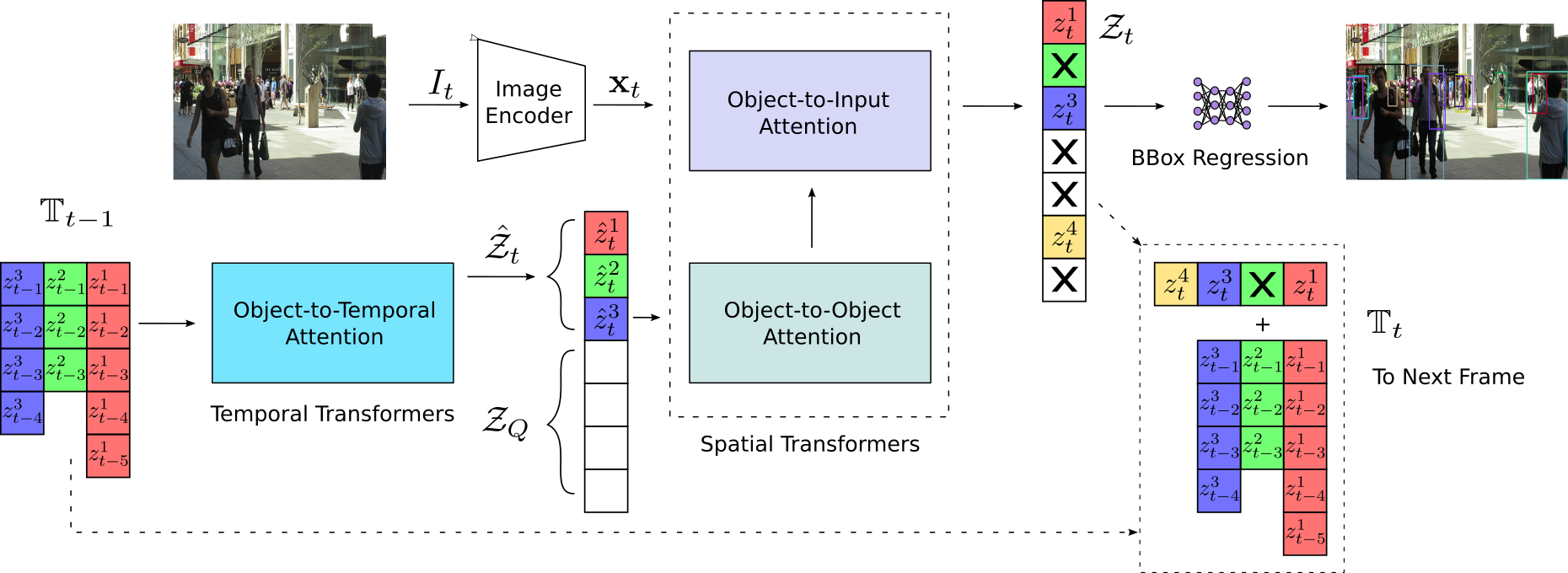}
}
\end{center}
\vspace{-.9em}
    \caption{Overview of our MO3TR framework. Starting from the left, the temporal Transformer uses the entire embedding-based track history~$\mathbb{T}_{t-1}$ to predict representative object encodings~$\hat{\mathcal{Z}}_t$ for the current, yet unobserved, time step~$t$. The spatial Transformer then jointly considers the predictions together with a set of learnt initiation embeddings~$\mathcal{Z}_Q$ and the input image~$I_t$ to reason about all objects in a joint manner, determining the initiation of new and termination of existing tracks. Embeddings of identified objects in~$\mathcal{Z}_t$ are used to regress corresponding bounding boxes describing the tracked objects, and are appended to form the track history~$\mathbb{T}_t$ for the next frame.}
    \vspace{-.5em}
\label{fig:System_Overview}
\end{figure*}

  
Precisely speaking, MO3TR incorporates long-term temporal information by casting \textit{temporal attention} over all past embeddings of each individual object, and uses this information to predict an embedding suited for the current time step (\textit{Object-to-Temporal Attention}, Fig.~\ref{fig:System_Overview}). This access to longer-term temporal information beyond two frames is crucial in enabling the network to learn the difference between occlusion and termination, which is further facilitated through a specific data augmentation strategy employed during training. To factor in the influence of other objects and the visual input measurements, we refine the predicted object embedding by casting spatial attention over all identified objects in the current frame (\textit{Object-to-Object Attention}) as well as over the objects and the encoded input image (\textit{Object-to-Input Attention}, Fig.~\ref{fig:System_Overview}).\par 
The idea of this joint approach relates to the natural way humans perceive such scenarios: We expect certain objects to become occluded given their past trajectory and their surroundings, and predict when and where they are likely to reappear.\par
To summarize, our main contributions are as follows:
\setlist{nolistsep}
\begin{enumerate}
	\setlength{\itemsep}{1pt}
	\setlength{\parskip}{0pt}
	\setlength{\parsep}{0pt}
    \item[1)] We introduce an end-to-end tracking approach that learns to encode longer-term information beyond two frames through temporal and spatial Transformers, and recursively predicts all states of the tracked objects\footnote{Code publicly available at \url{https://github.com/alanzty/MO3TR}}.
    \item[2)] We realize joint learning of object initialization, termination and occlusion handling without explicit data association and eliminate the need for heuristic post-processing.
    \item[3)] MO3TR reaches performances comparable or better than current state-of-the-art methods on several popular multi-object tracking benchmarks.
\end{enumerate}

\section{Related work}
\label{Related Work}
\noindent
\textbf{Tracking-by-detection.} Tracking-by-detection treats the multi-object tracking (MOT) task as a two-stage problem. Firstly, all objects in each frame are identified using an object detector~\cite{felzenszwalb2009object,ren2017accurate,NIPS2015_14bfa6bb,yang2016exploit}. Detected objects are then associated over frames, resulting in tracks~\cite{sort,chu2019online}.
The incorporation of appearance features and motion information has been proven to be of great importance for MOT.
Appearance and ReID features have been extensively utilized to improve the robustness of multi-object tracking~\cite{kim2015multiple,kuo2011does,leal2016learning,ristani2018features,yang2012online}. Further, incorporating motion has been achieved by utilizing a Kalman filter~\cite{kalman1960new} to approximate the displacement of boxes between frames in a linear fashion and with the constant velocity assumption~\cite{andriyenko2011multi,choi2010multiple} to associate detections~\cite{sort,deepsort}, or by incorporating environment information~\cite{particke2017improgpf} and possible intentions~\cite{particke2017multint,particke2018improvements} . Recently, more complex and data-driven models have been proposed to model motion~\cite{fang2018recurrent,liang2018lstm,zhang2020long,zhou2020centertrack} in a deterministic~\cite{ran2019robust,sadeghian2017tracking} and probabilistic~\cite{fang2018recurrent,saleh2020probabilistic,wan2018online} manner. 
Graph neural networks have been also used in the recent detection-based MOT frameworks, conducive to extract a reliable global feature representation from visual and/or motion cues ~\cite{braso2020learning,hornakova2020lifted,hornakova2021ldpnew,sheng2018heterogeneous,tang2017multiple,dai2021propGCN}, with many such methods utilising an offline approach to tackle the tracking problem.\par
Despite being highly interrelated, detection and tracking tasks are treated mostly as independent in this line of works. Further, the performance of tracking-by-detection methods often highly relies on incorporating heuristics and post-processing steps to infer track initiation and termination, handle occlusions and assign tracks.  
\par
\noindent\textbf{Joint detection and tracking.}
\revisedminor{The recent trend in MOT has been to move from disjoint detection and tracking components towards unified models, \eg by using the same network to jointly learn detection and embedding of objects~\cite{wang2020motjde}. While embeddings are here learnt explicitly with contrastive losses, our work learns the representations of the objects implicitly. Findings that the explicit learning of detection and appearance embeddings can often be contradicting have inspired approaches seeking to achieve fairness between the two tasks~\cite{zhang2021fairmot}. Other works have attempted to further improve joint detection and embedding learning through the structure of graph neural networks~\cite{wang2021joint}, or focused on regressing previous track locations to new locations in the current frame via a regression head~\cite{tracktor,feichtenhofer2017detect,zhou2020centertrack}.
Even deeper coupling of tracking and detection has recently been investigated by exploiting object motion cues from the previous frame(s) to better detect objects in the current frame~\cite{wu2021track}. This idea somewhat resembles the propagation of tracked objects in our framework which is one step towards learning tracking, but does not yet allow end-to-end learning of propagation, initialization and termination.
Advances in occlusion scenarios have been achieved in recent work by relying on the assumption that any physical object existing now will likely continue to exist during the period of tracking, helping to hallucinated the location of even fully occluded objects~\cite{tokmakov2021learning}.
Another recent stream of works is inspired by the concept of similarity learning that is dominant in many single generic object tracking methods~\cite{zhu2021learning} and provides strong discriminative power to distinguish objects of interest from their surroundings and can help propagation of existing objects~\cite{zheng2021improving,wang2021multiple,pang2021quasi}.
Although detection and tracking are not disjoint components in these works, they still suffer from some shortcomings. These works formulate the problem as detection matching between very few (mostly two) frames, thus solving the problem locally and ignoring long-term temporal information. Furthermore, these approaches still rely on conventional post processing steps and heuristics to generate the tracks, and recent gains have mainly been achieved due to specifically tailored components such as hand-designed track management modules~\cite{stadler2021tmoh}. 
Given the exhibited difficulties of most methods in increasingly complex tracking scenarios, we argue that solving the challenging MOT task requires long-term temporal encoding of object dynamics to handle object initiation, termination, occlusion and tracking -- a strategy that we take with our proposed method. 
}

\noindent\textbf{Transformers in vision.}
Recently, Transformers~\cite{attentionisallyouneed} have been widely applied to many computer vision problems~\cite{bello2019_attentionconv,carion2020detr, parmar2019_imagetransformer,rama2019_standalonetransf,khan2021transformers,dosovitskiy2021vit}, including two MOT methods~\cite{trackformer,sun2020transtrack} that have been developed concurrently to this work.
\cite{sun2020transtrack} performs multi-object tracking using a query-key mechanism which relies on heuristic post processing to generate final tracks. Trackformer~\cite{trackformer} has been proposed as a Transformer-based model which achieves joint detection and tracking by converting the existing DETR~\cite{carion2020detr} object detector to an end-to-end trainable MOT pipeline. However, it still considers solely local information (two consecutive frames) to learn and infer tracks and thus ignores long-term temporal object dynamics, which are essential for effective learning of all MOT components as we will demonstrate in the remainder of this work. \\
\noindent \textbf{This paper.} To overcome many of the aforementioned limitations of previous works, we propose an end-to-end multi-object tracking approach that leverages the architectural benefits of current Transformer models. Our method learns to jointly track multiple existing objects, handle their occlusion or terminate their tracks (depending on the situation), and to initiate new tracks -- all while considering long-term temporal object information.  

\section{MO3TR}
\label{MOTR}
Learning an object representation that encodes both the object's own state over time and the interaction with its surroundings is vital to allow reasoning about three key challenges present in end-to-end multi-object tracking (MOT), namely \textit{track initiation}, \textit{termination} and \textit{occlusion handling}. In this section, we demonstrate how such a representation can be acquired and continuously updated through our proposed framework: \underline{M}ulti-\underline{O}bject \underline{TR}acking using spatial \underline{TR}ansformers and temporal \underline{TR}ansformers -- short \textit{MO3TR}~(Fig.~\ref{fig:System_Overview}). We further introduce a training paradigm that facilitates learning of how to resolve these three challenges in a joint and completely end-to-end trainable manner. We start off by first presenting an overview of our framework and introduce the notation used throughout this paper, followed by a detailed introduction of the core components.\par

\subsection{System overview and notation}
\label{sec:system_overview}
The goal of tracking multiple objects in a video sequence of~$T$ frames is to retrieve an overall set of tracks~$\mathbb{T}_T$ representing the individual trajectories for all uniquely identified objects present in at least one frame of the sequence.
Given the first frame~$I_0$ at time~$t_0$, our model tentatively initializes a set of tracks~$\mathbb{T}_0$ based on all objects identified for this frame. 
From the next time step~$t>0$ onward, the model aims to compute a set of embeddings~$\mathcal{Z}_t$ = $\{\bm{z}^1_{t}, \bm{z}^2_{t}, ...\,, \bm{z}^M_{t}\}$ representing all~$M$ objects present in the scene at time~$t$ (Fig.~\ref{fig:System_Overview}). Taking in the track history~$\mathbb{T}_{t-1}$ from the previous time step, we predict a set of embeddings~$\hat{\mathcal{Z}}_t$ for the current time step based on the past representations of all objects using temporal attention (Fig.~\ref{fig:System_Overview}: \textit{Object-to-Temporal Attention}; Section~\ref{sec:temporal}). Together with a learnt set of representation queries~$\mathcal{Z}_Q$ proposing the initiation of new object tracks, these predicted object representations are processed by our first spatial attention module to reason about the interaction occurring between different objects (Fig.~\ref{fig:System_Overview}: \textit{Object-to-Object Attention}; Section~\ref{sec:spatial}). This refined set of intermediate object representations~$\mathcal{Z}'_t$ is then passed to the second spatial attention module which takes the interaction between the objects and the scene into account by casting attention over the object embeddings and the visual information of the current frame~$I_t$ transformed into its feature map~$\bm{x}_t$ (Fig.~\ref{fig:System_Overview}: \textit{Object-to-Input Attention}; Section~\ref{sec:spatial}). This two-step incorporation of spatial information into the embeddings is iteratively performed multiple times over several Transformer layers (alternating between self- and cross-attention), returning the final output set of refined object representations~$\mathcal{Z}_t$ for time step~$t$. Please refer to Section~\ref{sec:implementation} for architectural details. \par
The encoding of temporal and spatial information into a representative `state' embedding for any object~$m$ at time~$t$ 
\begin{equation}    
    \label{eq:repr_update_general}
    \bm{z}^{m}_{t} = f(\mathbb{T}_{t-1}, \mathcal{Z}_{Q}, \bm{x}_t) 
\end{equation}
can be summarized as a learnt function~$f(\cdot)$ of the track history~$\mathbb{T}_{t-1}$, the learnt set of initiation queries~$\mathcal{Z}_{Q}$ and the encoded image feature map~$\bm{x}_t$. This function representation demonstrates our main objective to enable the framework to learn the best possible way to relate the visual input to the objects' internal states, without enforcing overly-restrictive constraints or explicit data association.\par
The use of the resulting embedding set~$\mathcal{Z}_t$ in our framework is twofold. Tracking results in the form of object-specific class scores~$\bm{c}^{m}_{t}$ and corresponding bounding boxes $\bm{b}^{m}_{t}$ for the current frame are obtained through simple classification and bounding box regression networks (Fig.~\ref{fig:System_Overview}). Further, the subset of embeddings yielding a reasonably high probability of representing an object present in the current frame ($p_{\bm{z}_t^m}(c_{\text{obj}})\!>\!0.5$) is added to the track history to form the basis for the prediction performed in the next time step. Throughout the entire video sequence, new tracks~$\mathcal{T}^m_{s_m}$ representing objects that enter the scene are initialized,  while previous tracks may be terminated for objects no longer present. This leads to an overall set of tracks~$\mathbb{T}_T = \{\mathcal{T}^1_{s_1:e_1}, ...\,, \mathcal{T}^N_{s_N:e_N}\}$ for all $N$~uniquely identified objects present in at least one frame of the video sequence of length~$T$, with their life span indicated by the subscript as initiation time (\textit{start} `$s$') and termination time (\textit{end} `$e$').

\subsection{Learning long-term temporal embeddings}
\label{sec:temporal}
Discerning whether an object is not visible in a given frame due to occlusion or because it is no longer present in the scene can be challenging. Considering that visual features extracted at an object's expected position in the image during partial or full occlusion are not describing the actual object they aim to represent increases this even further. 
Humans naturally reach decisions in such scenarios by considering the available information jointly and over multiple time steps; we are able cherry-pick frames that provide helpful information, ignore the ones that prove unhelpful for the task, and predict \emph{if}, \emph{how} and \emph{where} an object is expected to re-appear in the current or a future frame. Intuitively, MO3TR follows a similar approach. \par
Our framework learns the temporal behavior of objects jointly with the rest of the model through a Transformer-based component~\cite{attentionisallyouneed} that we nickname \textit{temporal Transformer}. For any tracked object~$m$ at time~$t$, the temporal Transformer casts attention over all embeddings contained in the object's track history~$\mathcal{T}_{t-1}^{m}=\{\bm{z}^m_{s_m},...,\bm{z}^m_{t-1}\}$. Leveraging the information from previous frames encoded within these embeddings, it then predicts an \textit{expected} object representation~$\hat{\bm{z}}_t^m$ for the current time step~$t$. In other words, it predicts an object's state -- a condensed version of all the information required to reason about the object's most-likely location, identity and other properties to allow tracking.

While the architecture of a Transformer~\cite{attentionisallyouneed} is particularly suited for this purpose due to its ability to process and relate elements of sets like our track history, it is permutation-invariant in nature -- a property that is undesired in cases like ours where the indexing of the set plays a crucial role.  We thus supplement all embeddings~$\{\bm{z}^m_{s_m},...,\bm{z}^m_{t-1}\}$ that form part of an object's track history~$\mathcal{T}_{t-1}^{m}$ by adding positional encodings~\cite{attentionisallyouneed} to represent their relative time in the sequence and provide the Transformer in this way with the ability to leverage information from the time axis of the track. We denote the time-encoded track history by~$\mathcal{T}^{m,\text{pe}}_{t-1}$ and individual positional time encodings for time~$t$ as~$pe_t \in \mathbb{R}$. Passing the \textit{request} for an embedding estimate of the current time step~$t$ in form of the positional time encoding~$pe_t$ as a query to the Transformer\footnote{Note that this method allows to predict embeddings for any future time step, and could thus be easily extended to further applications like trajectory forecasting, or similar.}
and providing~$\mathcal{T}^{m,\text{pe}}_{t-1}$ as basis for keys and values, we retrieve the predicted object embedding
\begin{equation}
    \hat{\bm{z}}^{m}_{t} = \Psi\left(\frac{1}{\sqrt{d_{\bm{z}}}} \, q^{\text{tp}}\!\left(pe_t\right)\; k^{\text{tp}}\!\left(\mathcal{T}^{m,\text{pe}}_{t-1}\right)^{\text{T}} \right)\; v^{\text{tp}}\!\left(\mathcal{T}^{m,\text{pe}}_{t-1}\right),
    \label{eq:pred_temp}
\end{equation}
where~$\Psi$ represents the $\operatorname{softmax}$ operator, $q^{\text{tp}}(\cdot)$, $k^{\text{tp}}(\cdot)$ and $v^{\text{tp}}(\cdot)$ are learnt query, key and value functions of the temporal Transformer, respectively. The dimension of the object embeddings is denoted by $d_{\bm{z}}\in \mathbb{R}$.  \par
In other words, the predicted representation~$\hat{\bm{z}}_t^m$ of object~$m$ is computed through a dynamically weighted combination of all its previous embeddings. This allows the temporal Transformer to 
\begin{itemize}
    \item[(i)] incorporate helpful and ignore irrelevant or faulty information from previous time steps, and
    \item[(ii)] predict upcoming occlusions and create appropriate embeddings that might focus more on conveying important positional rather than visual information in such cases.
\end{itemize}
While these tasks resemble those usually performed via heuristics and manual parameter tuning during track management, MO3TR learns these dependencies end-to-end without the need of such heuristics.\par
In practice, the prediction of object representations introduced for the example of one tracked object in~\eqref{eq:pred_temp} is performed in a batched-parallel manner for the entire set of existing tracks~$\mathbb{T}_{t-1}$ over multiple Transformer layers, resulting in the output set~$\hat{\mathcal{Z}}_{t}$ of the temporal Transformers that is passed as input to the spatial Transformers~(Fig.~\ref{fig:System_Overview}). Note that the size of this set is dynamic and depends on the number of tracked objects. Details on how the temporal Transformer is trained are provided in Section~\ref{sec:training} and architectural details in Section~\ref{sec:implementation}.

\subsection{Learning spatial interactions}
\label{sec:spatial}
Multiple pedestrians that are present in the same environment not only significantly influence each others movements, but also their respective visual appearance through occluding each other when perceived from a fixed viewpoint. 
In this section, we introduce how MO3TR learns to incorporate these dependencies into the object representations. Starting from how detection and track initiation are performed within the concept of Transformers, we then detail the refinement of object embeddings by including the interaction between objects, followed by the interaction between objects and the input image. \par
\vspace{4pt}
\noindent\textbf{Initiation of new tracks.}
For a new and previously untracked object~$m$ spawning at any time~$t$, a corresponding track history~$\mathcal{T}^m_{t-1}$ does not yet exist and hence, no predicted embedding is passed from the temporal to the spatial Transformer~(Fig.~\ref{fig:System_Overview}). \revised{To allow initiation of new tracks for such objects, we build upon~\cite{carion2020detr} and learn a fixed set of initiation queries~$\mathcal{Z}_Q$ (in other words, a fixed set of learnt embeddings that acts as input to the decoder)}. Intuitively, these queries learn to propose embeddings that lead the spatial Transformer to check for objects with certain properties and at certain locations in the visual input data. Importantly, these queries are considered jointly with the ones propagated from the temporal Transformer to avoid duplicate tracks. Note that while our main method MO3TR is based around learning a fixed set of initiation queries, these can also be predicted for each image via a modified encoder -- an extension we discuss in Section~\ref{sec:extension_zq}.\par
\vspace{4pt}
\noindent\textbf{Interaction between tracked objects.} We use self-attention~\cite{attentionisallyouneed} to capture the influence tracked objects have onto each other's motion behavior and appearance. This interaction aspect is incorporated into the object embeddings by computing an updated version of the representation set
\begin{equation}
\label{eq:repr_spatial_update_1}
    \mathcal{Z}^{'}_t = \Psi\left(\frac{1}{\sqrt{d_{\bm{z}}}} \; q^{\text{sf}}(\bar{\mathcal{Z}}_t) \; k^{\text{sf}}(\bar{\mathcal{Z}}_t)^{\text{T}} \right) v^{\text{sf}}(\bar{\mathcal{Z}}_t),
\end{equation}
where $q^{\text{sf}}(\cdot)$, $k^{\text{sf}}(\cdot)$ and $v^{\text{sf}}(\cdot)$ are all learnt functions of the concatenated object embedding set~$\bar{\mathcal{Z}}_t=\{\hat{\mathcal{Z}}_t,\mathcal{Z}_Q\}$. The dimension of the embeddings is denoted by $d_{\bm{z}}$ and~$\Psi$ represents the~$\operatorname{softmax}$ operator. Relating this approach to the classical Transformer formulation, the learnt functions conceptually represent the queries, keys and values introduced in~\cite{attentionisallyouneed}, respectively.\par

\vspace{4pt}
\noindent\textbf{Interaction between objects and the input image.}
While the previous step of modeling the relationships between different objects can be interpreted as a refined way of proposing \textit{where to look} and \textit{what to look for} regarding existing and new objects, these proposals naturally have to be related to the actual input data of the current time step to draw conclusions about potential initiation, termination or further tracking of any given object. 
This relationship between the set of object embeddings and the input image is modeled through encoder-decoder attention (\textit{aka} cross-attention) to relate all object representations to the encoded visual information of the current image (\ie measurement). Evaluating this interaction results in the computation of a second update to the set of object representations
\begin{equation}
\label{eq:repr_spatial_update_2}
    \mathcal{Z}^{''}_{t} = \Psi\left(\frac{1}{\sqrt{d_{\bm{z}}}}\; q^{\text{cr}}(\mathcal{Z}'_t) \;k^{\text{cr}}(\bm{x}_t)^{\text{T}} \right) v^{\text{cr}}(\bm{x}_t),
\end{equation}
where $q^{\text{cr}}(\cdot)$ is a learnt function of the pre-refined object embeddings~$\mathcal{Z}'_t$, and $k^{\text{cr}}(\cdot)$ and $v^{\text{cr}}(\cdot)$ are learnt functions of the image embedding~$\bm{x}_t$ produced by a CNN backbone and a Transformer encoder. The embedding dimension is denoted by $d_{\bm{z}}$ and $\Psi$ represents the~$\operatorname{softmax}$ operator.\par
\vspace{4pt}
\noindent\textbf{Combining interactions for refined embeddings.} 
In practice, the two previously described update steps are performed consecutively with \eqref{eq:repr_spatial_update_2} taking as input the result of~\eqref{eq:repr_spatial_update_1}, and are iteratively repeated over several layers of the Transformer architecture. This sequential incorporation of updates into the representation is inspired by DETR~\cite{carion2020detr}, where self-attention and cross-attention modules are similarly deployed in a sequential manner. Using both introduced concepts of \textit{Object-to-Object} and \textit{Object-to-Input} attention allows our model to globally reason about all tracked objects via their pair-wise relationships, while employing the current image as context information to retrieve the final set of updated object representations~$\mathcal{Z}_t$.\par
\vspace{4pt}
\noindent\textbf{Updating the track history.} After each frame is processed by the entire framework, the final set of embeddings~$\mathcal{Z}_t$ of objects identified to be present in the frame is added to the track history~$\mathbb{T}_{t-1}$, creating the basis for the next prediction of embeddings by the temporal Transformer~(Fig.~\ref{fig:System_Overview}). We consistently append new embeddings from the right-hand side, followed by right-aligning the entire set of embeddings. Due to the different lengths of tracks for different objects, this procedure aligns embeddings representing identical time steps, a method that we found to help stabilize training and improve the inference of the temporal Transformer (see ablation studies in Section~\ref{sec:ablations} and Table~\ref{tab: training}). 

\subsection{Removing the need for explicit data association}
\label{sec:dataassoc}
In the introductory section of this paper, we briefly noted that our proposed MO3TR performs tracking \textit{without any explicit data association module}.
To elaborate what we mean by that and how multi-object tracking (MOT) without an explicitly formulated data association task is feasible, let us reconsider the actual definition of the MOT problem: Finding a mapping from any given input data, \eg an image sequence, to the output data, \ie a set of object states over time. In any learning scheme, given a suitable learning model, this mapping function can theoretically be learned end-to-end without the requirement for solving any additional auxiliary task, as long as the provided inputs and outputs are clearly defined. The firmly established task of performing data association between detections and objects, usually via a minimum cost assignment solved by using the Hungarian algorithm, is nothing more than such an auxiliary task that has originally been created to solve tracking based on the tracking-by-detection paradigm. An end-to-end learning model, however, can learn to directly infer implicit correspondences -- thus rendering the explicit formulation of this task obsolete. \par 
Precisely speaking, our end-to-end tracking model learns to relate the visual input information to the internal states of the objects via a self-supervised attention scheme. We realize this through the introduced combination of Transformers~\cite{attentionisallyouneed} to distill the available spatial and temporal information into representative object embeddings (\ie the object states), making the explicit formulation of any auxiliary data association strategy unnecessary.

\subsection{Training MO3TR}
\label{sec:training}
The training procedure of MO3TR (Fig.~\ref{fig:System_Overview}) is composed of two key tasks: (i) creating a set of suitable tracklets that can be used as input~$\mathbb{T}_{t-1}$ to the temporal Transformer, and (ii) assigning the predicted set of~$M$ output embeddings~$\mathcal{Z}_t=\{\bm{z}_t^m\}_{m=1}^M$ to corresponding ground truth labels of the training set, and applying a corresponding loss to facilitate training. With the number of output embeddings being by design larger than the number of objects in the scene, matching occurs either with trackable objects or the background (no-object) class.\par
\vspace{4pt}
\noindent \textbf{Constructing the input tracklet set.} The input to the model at any given time~$t$ is defined as the track history~$\mathbb{T}_{t-1}$ and the current image $I_t$. To construct a corresponding~$\mathbb{T}_{t-1}$ for any~$I_t$ sampled from the dataset during training, we first extract the ordered set of~$K$ directly preceding images~$\{I_{k}\}_{k=t-K}^{t-1}$ from the training sequence. Passing these images without track history to MO3TR causes the framework to perform track initiation for all identified objects in each frame by using the trainable embeddings~$\mathcal{Z}_Q$, returning an ordered set of output embedding sets~$\{\mathcal{Z}_k\}_{k=t-K}^{t-1}$. Each output embedding set~$\mathcal{Z}_k$ contains a variable number of $M_k$~embeddings representing objects in the respective frame~$k$. We use multilayer perceptrons (MLPs) to extract corresponding bounding boxes $\hat{\bm{b}}_k^m$ and class scores $\hat{\bm{c}}_k^m$ from each of these object embeddings~$\bm{z}^m_k\in\mathcal{Z}_k$, resulting in a set of~$M_k$ object-specific pairs denoted as~$\{\hat{y}^m_k\}_{m=1}^{M_k}\!=\!\{(\hat{\bm{b}}_k^m, \hat{\bm{c}}_k^m)\}_{m=1}^{M_k}$ for each frame~$k$. The pairs are then matched with the ground truth~$\{y^i_k\}_{i=1}^{G_k}$ of the respective frame through computing a bipartite matching between these sets following~\cite{carion2020detr}. The permutation~$\hat{\sigma}_k$ of the~$M_k$ predicted elements with lowest pair-wise matching cost~$\mathcal{C}_{\text{matching}}$ is determined by solving the assignment problem 
\begin{equation}
    \hat{\sigma}_k = \operatornamewithlimits{arg\,min}_{\sigma\in\mathscr{S}}\sum_{i}^{M_k} \mathcal{C}_{\text{matching}}\!\left(y_k^i,\hat{y}_k^{\sigma(i)}\right),
    \label{eq:permute}
\end{equation}
through the Hungarian algorithm~\cite{kuhn1955hungarian}, with the matching cost taking both the probability of correct class prediction~$\hat{p}_k^{\sigma(i)}(\bm{c}_k^i)$ and bounding box similarity into account 
\begin{equation}
    \mathcal{C}_{\text{matching}} = -\hat{p}_k^{\sigma(i)}\left(\bm{c}_k^i\right) + \mathcal{C}_{\text{bbox}}\!\left(\bm{b}_k^i,\bm{\hat{b}}_k^{\sigma(i)}\right).
    \label{eq:matchcost}
\end{equation}
We follow~\cite{carion2020detr} and use a linear combination of L1 distance and the scale-invariant generalized intersection over union~\cite{giou} cost~$\mathcal{C}_\text{giou}$ to mitigate any possible scale issues arising from different box sizes. The resulting bounding box cost with weights~$\alpha_{\text{L1}},\alpha_{\text{giou}} \in \mathbb{R}^+$ is then defined as
\begin{equation}
    \mathcal{C}_{\text{bbox}} = \alpha_{\text{L1}} \left\lVert \bm{b}^i_k - \bm{\hat{b}}_k^{\sigma(i)} \right\rVert_1 + \alpha_{\text{giou}}\mathcal{C}_{\text{giou}}\left(\bm{b}^i_k,\bm{\hat{b}}_k^{\sigma(i)}\right).
    \label{eq:boxcost}
\end{equation}
The identified minimum cost matching between the output and ground truth sets is used to assign all embeddings classified as objects their respective identities annotated in the ground truth labels. The objects of all~$K$ frames are accumulated, grouped regarding their assigned identities and sorted in time-ascending order to form the overall set of previous object tracks~$\mathbb{T}_{t-1}$ serving as input to our model.\par
\vspace{4pt}
\noindent\revised{\textbf{Beyond two frames.} Most recent tracking works attempt \ignore{employ a two-frame concept }to learn track initiation and termination by sampling two frames out of a short sequence. This is however only a sparse and often poor approximation of the actual complex tracking scenario and thus frequently requires significant post-processing or dataset specific tuning of the augmentation process (\eg adjusting the false positive rate) due to the gap between the scenarios encountered during training and at inference time. In this work, we attempt to take a step towards closing this gap and use our model to infer the results for up to 30 consecutive frames during training. We then employ the same algorithm underlying the CLEAR metric~\cite{bernardin2008evaluating} to match our predicted tracklets and ground truth annotations. This procedure better replicates the scenarios encountered during inference already during the training process, allowing for complex scenarios to naturally unfold and for errors to accumulate over multiple frames -- thus helping to learn the characteristics of the tracking problem from the data itself rather than based on heuristic choices of dataset-specific augmentation parameters.}\par
\vspace{4pt}
\noindent \textbf{Losses.} Given the created input set of tracks~$\mathbb{T}_{t-1}$ and the image $I_t$, MO3TR predicts an output set of object embeddings~$\mathcal{Z}_{t}$ = $\{\bm{z}^1_{t}, \bm{z}^2_{t}, ...\,, \bm{z}^M_{t}\}$ at time~$t$. Similar to before, we extract bounding boxes and class scores for each embedding in the set. However, embeddings that possess a track history already have unique identities associated to them and are thus directly matched with the respective ground truth elements. Only newly initiated  embeddings without track history are then matched with remaining unassigned ground truth labels as previously described. Elements that could not be matched are assigned the \textit{background} class. Finally, we re-use~\eqref{eq:matchcost} and~\eqref{eq:boxcost} for~$k=t$ and apply them as our loss to the matched elements of the output set. 
\par
\vspace{4pt}
\noindent \textbf{Data augmentation.}
Most datasets are highly imbalanced regarding the occurrence of occlusion, initiation and termination scenarios. To facilitate learning of correct tracking behaviour and aide our previously described multi-frame training procedure, we propose to further mitigate the imbalance problem by modelling similar effects through augmentation:
\begin{enumerate}
    \item We randomly drop a certain number of embeddings in the track history to simulate cases where the object could not be identified for some frames, aiming to increase robustness. If the most recent embedding is dropped, the model can learn to re-identify objects. 
    \item Random false positive examples are inserted into the history to simulate false detection and faulty appearance information due to occlusion. This aims at encouraging the model to learn to ignore unsuited representations through its attention mechanism.
    \item We randomly select the sequence length used to create the track history during training to increase the model's capability to deal with varying track lengths.
\end{enumerate}
The importance of these augmentation strategies is investigated and clearly demonstrated during our ablation studies in Section~\ref{sec:ablations} and Table~\ref{tab: training}, leading to significant improvements compared to non-augmented training.

\section{Experiments and discussion}
\label{Experiments}
\begin{table*}
    \begin{center}
    \caption{Results on the MOT16 benchmark~\cite{mot17} test set using \textbf{public} detections. \textbf{Bold} and \underline{underlined} numbers indicate best and second best result, respectively. In case the two versions of our method fill both spots, we additionally use \dotuline{dotted} underline to indicate the next best result. Metrics that were not available for comparison are indicated via a hyphen~(--). More detailed results of our approach across the individual sequences of the benchmark are provided in the supplementary material. $^\dagger$ MT and ML in~\cite{tian2019detection} were reported without fractional digit.}
    \label{tab:MOT16}
    \scalebox{0.95}
    {
    {
    \setlength{\tabcolsep}{6pt}
    \begin{tabular}{@{}c l *5c r r c@{}} 
    \toprule
         &  Method  & {MOTA$\uparrow$} & {IDF1$\uparrow$} & {HOTA$\uparrow$}  & {MT$\uparrow$} & {ML$\downarrow$} & {FP$\downarrow$}  & {FN$\downarrow$} & {IDs$\downarrow$}\\    \midrule 
        \multirow{6}{*}{\STAB{\rotatebox[origin=c]{90}{Offline}}} 
        & eHAF~\cite{sheng2018heterogeneous} & 47.2 & 52.4 & 40.3 & 18.6 & 42.8 & 12,586 & 83,107 & 542\\
        & NOTA~\cite{chen2019aggregate} & 49.8 & 55.3 & 40.7 & 17.9 & 37.7 & 7,428 & 83,614 & 614\\
        & MPNTrack~\cite{braso2020learning} & 58.6 & 61.7 & 48.9 & \underline{27.3} & \underline{34.0} & \underline{4,949} & 70,252 & \textbf{354}\\
        & LPC\_MOT~\cite{dai2021propGCN} & 58.8 & \textbf{67.6} & \textbf{51.7} & \underline{27.3} & 35.0 & 6,167 & 68,432 & 435\\
        & Lif\_T~\cite{hornakova2020lifted} & \underline{61.3} & 64.7 & 50.8 & 27.0 & \underline{34.0} & \textbf{4,844} & \underline{65{,}401} & \underline{389}\\
        & ApLift~\cite{hornakova2021ldpnew} & \textbf{61.7} & \underline{66.1} & \underline{51.3} & \textbf{34.3} & \textbf{31.2} & 9{,}168 & \textbf{60,180} & 495\\
        \midrule
        \multirow{13}{*}{\STAB{\rotatebox[origin=c]{90}{Online}}}
        & EAMTT~\cite{sanchez2016_onlinemot} & 38.8 & 42.4 & 32.5 & 7.9 & 49.1 & 8,114 & 102,452 & 965\\
        & DMAN~\cite{zhu2018_DMAN} & 46.1 & 54.8 & 40.3 & 17.4 & 42.7 & 7,909 & 89,874 & \textbf{532}\\
        & AMIR~\cite{sadeghian2017tracking} & 47.2 & 46.3 & 35.1 & 14.0 & 41.6 & \underline{2{,}681} & 92,856 & 774\\
        & MOTDT17~\cite{long2018_tracking} & 47.6 & 50.9 & 38.4 & 15.2 & 38.3 & 9,253 & 85,431 & 792\\
        & STRN~\cite{xu2019spatial} & 48.5 & 53.9 & 39.7 & 17.0 & 34.9 & 9,038 & 84,178 & 747\\
        & UMA~\cite{yin2020unified} & 50.5 & 52.8 & -- & 17.8 & 33.7 & 7,587 & 81,924 & 685\\
        & Tracktor++~\cite{tracktor} & 54.4 & 52.5 & 42.3 & 19.0 & 36.9 & 3,280 & 79,149 & 682\\
        & DeepMOT-T~\cite{xu2020train} & 54.8 & 53.4 & 42.2 & 19.1 & 37.0 & 2,955 & 78,765 & 645\\
        & Tracktor++v2~\cite{tracktor} & 56.2 & 54.9 & 44.6 & 20.7 & 35.8 & \textbf{2{,}394} & 76{,}844 & \underline{617}\\
        & TINF~\cite{tian2019detection}$^{\dagger}$ & 57.6 & -- & -- & \underline{30.--} & \underline{22.--} & 12.121 & 64,401 & 733\\
        & TMOH~\cite{stadler2021tmoh} & \dotuline{63.2} & \textbf{63.5} & \textbf{50.7} & 27.0 & 31.0 & 3,122 & \dotuline{63{,}376} & 635\\
        \cmidrule{2-10}
        & MO3TR (Res50, IoU) & \underline{63.5} & 60.3 & 49.6 & 27.4 & \underline{22.4} & 6,987 & \underline{58{,}691}  & 880  \\ 
        & MO3TR (Res50, CD) & \textbf{64.2} & \underline{60.6} & \underline{50.3} & \textbf{31.6} & \textbf{18.3} & 7,620 & \textbf{56{,}761}  & 929  \\
        \bottomrule
    \end{tabular}
    }
    }
    \end{center}
\end{table*}
In this section, we demonstrate and discuss the performance of MO3TR by comparing against other multi-object tracking methods on popular MOT benchmarks\footnote{\url{https://motchallenge.net/}} and evaluate different aspects of our contribution in detailed ablation studies. We further provide all implementation details required to reproduce our findings.\par
\vspace{4pt}
\noindent \textbf{Datasets.} We use the widely established MOT16 and MOT17~\cite{mot17} datasets as well as the rather new MOT20 dataset~\cite{mot20} from the MOTchallenge benchmarks to evaluate and compare MO3TR with other state of the art models. Both MOT16 and MOT17 contain seven training and test sequences each, capturing crowded indoor or outdoor areas via moving and static cameras from various viewpoints. MOT20 contains eight sequences (four training and four test), focusing on heavily crowded scenes with a very high number of people in both day and night scenarios. Pedestrians are often heavily occluded by other pedestrians or background objects across all three datasets, making identity-preserving tracking challenging. Three sets of public detections are provided with MOT17 (DPM~\cite{felzenszwalb2009object}, FRCNN~\cite{NIPS2015_14bfa6bb} and SDP~\cite{yang2016exploit}), one with MOT16 (DPM) and one with MOT20 (FRCNN). 
For ablation studies, we combine sequences of the new MOT20 benchmark~\cite{mot20} and 2DMOT15~\cite{2dmot15} to form a diverse validation set covering both indoor and outdoor scenes at various pedestrian density levels. \par
\vspace{4pt}
\noindent \textbf{Evaluation metrics.} 
\label{sec:eval_metrics}
To evaluate MO3TR and compare its performance to other state-of-the-art tracking approaches, we use the standard set of metrics recognized by the tracking community~\cite{bernardin2008evaluating,ristani2016performance}. Analyzing the detection performance, we provide detailed insights regarding the total number of \textit{false positives}~(FP) and \textit{false negatives}~(FN, \ie missed targets). The \textit{mostly tracked targets}~(MT) measure describes the ratio of ground-truth trajectories that are covered for at least 80\% of the track's life span, while \textit{mostly lost targets}~(ML) represents the ones covered for at most 20\%. The number of \textit{identity switches} is denoted by IDs. The two most commonly used metrics to summarize the tracking performance are the \textit{multiple object tracking accuracy} (MOTA), and the identity F1 score (IDF1). MOTA combines the measures for the three error sources of false positives, false negatives and identity switches into one compact measure, and a higher MOTA score implies better performance of the respective tracking approach. The IDF1 represents the ratio of correctly identified detections over the average number of ground-truth and overall computed detections. We additionally report our results on the recently proposed \textit{higher order tracking accuracy} metric (HOTA), which aims to combine the effects of accurate detection, association and localization, and better aligns with human perception of tracking performance~\cite{luiten2021hota}. \par
All reported results are computed using the official evaluation code of the MOTChallenge benchmark. For further details, please refer to the supplementary material.

\subsection{Implementation details of MO3TR}
\label{sec:implementation}
\noindent \textbf{Multi-stage training details.} 
\revised{To implement the general training strategy introduced in Section~\ref{sec:training}, we employ a multi-stage training concept to train MO3TR end-to-end. Firstly, our COCO~\cite{lin2014coco} pretrained ResNet50~\cite{he2016resnet} backbone is, together with the encoder and spatial Transformers, trained on the combination of the CrowdHuman~\cite{crowdhuman}, ETH~\cite{ethdataset} and CUHK-SYSU~\cite{xiao2017joint} datasets for 300 epochs on a pedestrian detection task. This training procedure is similar to the one introduced in DETR~\cite{carion2020detr}. Afterwards, we engage our temporal Transformer and train the entire model end-to-end using the respective dataset (e.g. MOT17) for another 100 epochs. The initial learning rate for both training tasks is 1e-4, and is dropped by a factor of 10 at $0.3$ and $0.4$ times the total number of epochs (\ie at 90$|$120 and 60$|$80 for first and second training stage, respectively).} Relative weights of our loss are the same as described in~\cite{carion2020detr}, and we choose the number of initiation queries to~$\mathcal{Z}_Q=100$. The input sequence length representing object track histories~len($\mathcal{T}^m_{t-1}$) for any object~$m$ varies randomly from 1 to 30 frames \revised{-- meaning that we can expect our model to re-identify objects that have been occluded for at most this number of frames. To reduce overall memory consumption, we thus drop the tracklet for any object~$m$ from our set of active tracks~$\mathbb{T}$ if the respective object has not been successfully identified for 30 frames. Note that this cut-off is a practical design choice given our limited computational resources, and does not affect the capability of our approach to track across longer sequences.} To enhance the learning of temporal encoding, we predict 10 future frames instead of one and compute the total loss. \revised{Please note that while we choose a multi-stage training strategy for increased training stability and reduced initial memory footprint, our approach is fully differentiable without the use of non-differentiable heuristics and can thus be trained entirely `end-to-end'.} \revisedminor{We provide additional experiments and discussion on using a single-stage joint training approach in the supplementary material.} \par
\vspace{4pt}
\noindent \textbf{Model components.}
We use an 8-head 3-layer self-attention module as temporal Transformer (Section~\ref{sec:temporal}) and 6 pairs of 8-head 6-layer modules performing self- and cross-attention in an alternating manner as spatial Transformer (Section~\ref{sec:spatial}). Both Transformers use 256d embeddings, and we use 1D and 2D sinusoidal positional encodings to supplement the input data for the temporal and spatial Transformers, respectively. Classification and bounding box regression are performed via one and three layer MLPs. \par
\vspace{4pt}
\noindent \textbf{Computational complexity for training.} 
\revised{We train our MO3TR model using 4 GTX 1080ti GPUs with 11GB memory each with a total batch size of 32 during the first training stage (detection task), and on one single GPU with a batch size of 4 in the second stage.} It is to be noted that these computational requirements are significantly lower than for other recently published approaches in this field. We expect the performance of our model to further increase through bigger backbones and longer sequence lengths as well as an increased number of objects per frame. \par
\vspace{4pt}
\noindent \textbf{Public detection.} 
We evaluate the tracking performance using the public detections provided by the MOTChallenge. Since our method is based on embeddings and thus not able to directly produce tracks from these provided detections, we follow~\cite{zhou2020centertrack} in filtering our initiations by the public detections using bounding box center distances (CD), and only allow initiation of matched and thus publicly detected tracks. We additionally report our results achieved by using the strategy of using the public detections to filter track initiations via the Intersection over Union (IoU) recently introduced in~\cite{trackformer}.\par
\vspace{4pt}
\noindent \textbf{Private detection.} 
To facilitate better comparison to other methods, we additionally report the performance on tracking benchmark using our method's own, unfiltered detections -- commonly referred to as \textit{private} detections. 

\subsection{Extending MO3TR by predicting initiation queries}
\label{sec:extension_zq}
\revised{Observing that current set-based detectors like DETR~\cite{carion2020detr} with ResNet~\cite{he2016resnet} backbone are still often significantly outperformed by non-set based versions like~\cite{ge2021yolox}, we explore the possibility of extending our MO3TR setup via the use of a more powerful multi-scale encoder to improve the detection capabilities of our framework. For our investigations, we choose the multi-scale Deformable DETR~\cite{zhu2020deformable} with the multi-scale DarkNet CNN backbone~\cite{ge2021yolox} (initialized with COCO-pretrained~\cite{lin2014coco} weights) and train this new encoder on the CrowdHuman~\cite{crowdhuman} and MOT20~\cite{mot17} datasets using the detection head and losses proposed in~\cite{ge2021yolox}, following their training strategy and hyperparameter setup. Given our computational resource constraints, we then freeze the backbone and head, and continue to train our tracking framework for 50 epochs on the respective MOT dataset. Instead of learning a fixed set of initiation queries like with our base MO3TR version, we investigate the advantage of dynamic query generation via the encoder and directly retrieve our desired set of initiation queries~$\mathcal{Z}_Q$ from the encoder. In detail, we use the classifier part of the detection head to select the top $k=100$ embeddings from the Darknet output as initiation queries, concatenate them as previously introduced with the output set of our temporal Transformer and pass it to the spatial Transformer module as described in Sections~\ref{sec:temporal} and~\ref{sec:spatial}. We refer to this extended method as MO3TR-PIQ (due to its \underline{P}redicted \underline{I}nitiation \underline{Q}ueries).}

\subsection{Comparison with the state of the art}

\begin{table*}
    \begin{center}
    \caption{Results on the MOT17 benchmark~\cite{mot17} test set using \textbf{public} detections. \textbf{Bold} and \underline{underlined} numbers indicate best and second best result, respectively. Metrics that were not available for comparison are indicated via a hyphen~(--). Public detections (PD) are processed in different ways by different methods based on their algorithmic structure: directly as input (IN), or used for filtering of track initiation proposals via the center distance (CD) as proposed in~\cite{zhou2020centertrack} or the intersection over union (IoU)~as proposed in~\cite{trackformer}. More detailed results of our approach are provided in the supplementary material. $^\dagger$~Result reported in~\cite{trackformer}.}
    \vspace{-.5em}
    \label{tab:MOT17}
    {
    {
    \setlength{\tabcolsep}{3pt}
    \setlength{\tabcolsep}{6pt}
    \begin{tabular}{@{}c *2l *2c *7r@{}} 
    \toprule
          & Method  & Backbone & PD & {MOTA$\uparrow$}  & {IDF1$\uparrow$} & {HOTA$\uparrow$}  & {MT$\uparrow$} & {ML$\downarrow$} & {FP$\downarrow$}  & {FN$\downarrow$} & {IDs$\downarrow$}\\    \midrule 
          \multirow{9}{*}{\STAB{\rotatebox[origin=c]{90}{Offline}}} 
          & jCC~\cite{keuper2018motion}& -- & IN & 51.2 & 54.5 & 42.5 & 20.9 & 37.0  & 25,937 & 247,822 & 1{,}802\\
          & NOTA~\cite{chen2019aggregate} & -- & IN & 51.3 & 54.7 & 42.3 & 17.1 & 35.4 & 20,148 & 252,531 & 2,285\\
          & eHAF~\cite{sheng2018heterogeneous} & -- & IN & 51.8 & 54.7 & 43.4 & 23.4 & 37.9 & 33,212 & 236,772 & 1,834\\
          & JBNOT~\cite{henschel2019multiple} & -- & IN & 52.6 & 50.8 & 41.3 & 19.7 & 35.8 & 31,572 & 232,659 & 3,050\\
          & TT~\cite{zhang2020long} & -- & IN & 54.9 & 63.1 & 48.4 & 24.4 & 38.1 & 20{,}236 & 233,295 & \textbf{1,088}\\
          & MPNTrack~\cite{braso2020learning} & -- & IN & 58.8 & 61.7 & 49.0 & 28.8 & \underline{33.5} & \underline{17,413} & 213,594 & 1,185\\
          & LPC\_MOT~\cite{dai2021propGCN} & -- & IN & 59.0 & \textbf{66.8} & \textbf{51.5} & \underline{29.9} & 33.9 & 23,102 & 206,948 & \underline{1,122}\\
          & Lif\_T~\cite{hornakova2020lifted} & -- & IN & \textbf{60.5} & \underline{65.6} & \underline{51.3} & 27.0 & 33.6 & \textbf{14{,}966} & \underline{206,619} & 1,189\\
          & ApLift~\cite{hornakova2021ldpnew} & -- & IN & \textbf{60.5} & \underline{65.6} & 51.1 & \textbf{33.9} & \textbf{30.9} & 30{,}609 & \textbf{190,670} & 1,709\\
          \midrule[0.7pt]
          
        \multirow{20}{*}{\STAB{\rotatebox[origin=c]{90}{Online}}}
        & GMPHD~\cite{kutschbach2017sequential} & -- & IN & 39.6 & 36.6 & 30.3 & 8.8 & 43.3 & 50,903 & 284,228 & 5,811\\
        & EAMTT~\cite{sanchez2016_onlinemot} & ResNet50~\cite{he2016resnet} & IN & 42.6 & 41.8 & -- & 12.7 & 42.7 & 30,711 & 288,474 & 4,488\\
        & SORT17~\cite{sort}& VGG16~\cite{simonyan2014very} & IN & 43.1 & 39.8 & 34.0 & 12.5 & 42.3 & 28,398 & 287,582 & 4,852\\
        & DMAN~\cite{zhu2018_DMAN}& ResNet50~\cite{he2016resnet} & IN & 48.2 & \underline{55.7} & 42.5 &  19.3 & 38.3 & 26,218 & 263,608 & 2,194\\
        & MOTDT17~\cite{long2018_tracking}& GoogLeNet~\cite{szegedy2015going} & IN & 50.9 & 52.7 & 41.2 & 17.5 & 35.7 & 24,069 & 250,768 & 2,474\\
        & STRN~\cite{xu2019spatial}& ResNet50~\cite{he2016resnet} & IN & 50.9 & \textbf{56.5} & \underline{42.6} & 20.1 & 37.0 & 27,532 & \underline{246,924} & 2,593\\
        & DeepMOT-T~\cite{xu2020train}& ResNet101~\cite{he2016resnet} & IN & 53.7 & 53.8 & 42.4 & 19.4 & 36.6 & \underline{11,731} & 247,447 & \textbf{1{,}947}\\
        & FAMNet~\cite{chu2019famnet}& ResNet101~\cite{he2016resnet} & IN & 52.0 & 48.7 & -- & 19.1 & \underline{33.4} & 14,138 & 253,616 & 3,072\\
        & UMA~\cite{yin2020unified}& AlexNet~\cite{krizhevsky2012imagenet} & IN & 53.1 & 54.4 & -- & \textbf{21.5} & \textbf{31.8} & 22,893 & \textbf{239{,}534} & 2,251\\
        & Tracktor++~\cite{tracktor}&ResNet101~\cite{he2016resnet} & IN & \underline{53.5} & 52.3 & 42.1 & 19.5 & 36.6 & 12,201 & 248,047 & \underline{2,072}\\
        & Tracktor++v2~\cite{tracktor}&ResNet101~\cite{he2016resnet} & IN & \textbf{56.5} & 55.1 & \textbf{44.8} & \underline{21.1} & 35.3 & \textbf{8{,}866} & 248,047 & 3,763\\
        \cmidrule{2-12}
       & CenterTrack \cite{zhou2020centertrack}$^\dagger$ & DLA~\cite{yu2018dla}& IoU & 60.5 & 55.7 & -- & 26.4 & 33.0 & \textbf{11,599} & 208,577 & \textbf{2{,}540}\\
        & TrackformerV1 \cite{trackformer} & ResNet101~\cite{he2016resnet} & IoU & 59.7 & \underline{59.0} & -- & \textbf{29.6} & \underline{25.1} & 30,724 & 194{,}320  & \underline{2{,}579}\\
        & \revised{TrackformerV3 \cite{trackformer}} & \revised{ResNet50~\cite{he2016resnet}} & \revised{IoU} & \revised{\underline{62.3}} & \revised{57.6} & \revised{--} & \revised{\underline{29.2}} & \revised{27.1} & \revised{\underline{16,591}} & \revised{\underline{192{,}123}}  & \revised{4,018}\\
        & MO3TR (ours)  & ResNet50~\cite{he2016resnet} & IoU & \textbf{62.7} & \textbf{59.9} & \textbf{49.4} & 28.4 & \textbf{23.2} & {20,075} & \textbf{187,578} & 2,680 \\
        \cmidrule{2-12}
        & CenterTrack \cite{zhou2020centertrack}& DLA~\cite{yu2018dla} & CD & 61.5 & 59.6 & \underline{48.2} & 26.4 & 31.9 & \textbf{14,076} & 200,672 & \textbf{2,583}\\
        & TrackformerV1 \cite{trackformer} & ResNet101~\cite{he2016resnet} & CD & 61.8 & 59.8 & -- & \textbf{35.4} & \underline{21.1} & 35,226 & \textbf{177{,}270}  & 2,982\\
        & \revised{TrackformerV3~\cite{trackformer}} & \revised{ResNet50~\cite{he2016resnet}} & \revised{CD} & \revised{\textbf{63.4}} & \revised{\underline{60.0}} & -- & {--} & -- & -- & --  & --\\
        & MO3TR (ours)  & ResNet50~\cite{he2016resnet} & CD & \underline{63.2} & \textbf{60.2} & \textbf{49.6} & \underline{31.9} & \textbf{19.2} & \underline{21,966} & \underline{182,860}  & \underline{2,841} \\
        \bottomrule
    \end{tabular}
    }
    }
    \end{center}
    \vspace{-.5em}
\end{table*}

\begin{table}
    \begin{center}
    \vspace{-.5em}
    \caption{Results on the MOT17 benchmark~\cite{mot17} test set using \textbf{private} detections. \textbf{Bold} and \underline{underlined} numbers indicate best and second best result, respectively. More detailed results of our approach are provided in the supplementary material.}
    \vspace{-.5em}
    \label{tab: MOT17Private}
    \scalebox{0.95}
    {
    {
    \setlength{\tabcolsep}{3pt}
    \begin{tabular}{@{}l *3c r r c@{}} 
    \toprule
          Method  & {MOTA$\uparrow$} & {IDF1$\uparrow$} & {HOTA$\uparrow$} & {FP$\downarrow$}  & {FN$\downarrow$} & {IDs$\downarrow$}\\    \midrule  
        TubeTK~\cite{pang2020tube} & 63.0 & 58.6 & 48.0 & 27,060 & 177,483 & 4,137\\
        CTracker~\cite{peng2020chained} & 66.6 & 57.4 & 52.2 &  22,284 & 160,491 & 5,529\\
        CenterTrack~\cite{zhou2020centertrack} & 67.8 & 64.7 & 49.0 &  \textbf{18,498} & 160,332 & 3{,}039\\
        TrackformerV2~\cite{trackformer} & 65.0 & 63.9 &-- & 70,443 & 123{,}552 & 3,528\\
        \revised{TrackformerV3~\cite{trackformer}} & \revised{74.1} & \revised{68.0} & -- & \revised{34,602} & \revised{108{,}777} & \revised{\textbf{2,829}}\\
        \revised{TraDeS~\cite{wu2021track}} & \revised{69.1} & \revised{63.9} & \revised{52.7} & \revised{20,892} & \revised{150,060}& \revised{3,555}\\
        \revised{PermaTrack~\cite{tokmakov2021learning}}& \revised{73.8}& \revised{68.9}& \revised{55.5}& \revised{28,998}& \revised{115,104}& \revised{3,699}\\
        \revised{GSDT-V2~\cite{wang2021joint}}& \revised{73.2}& \revised{66.5}& \revised{55.2}& \revised{26,397}& \revised{120,666}& \revised{3,891}\\
        \revised{SOTMOT~\cite{zheng2021improving}}& \revised{71.0}& \revised{71.9}& \revised{--}& \revised{39,537}& \revised{118,983}& \revised{5,189}\\
        \revised{CorrTracker~\cite{wang2021multiple}}& \revised{\underline{76.5}}& \revised{\textbf{73.6}}& \revised{--}& \revised{29,808}& \revised{\textbf{99,510}}& \revised{3,369}\\
        \revised{FairMOTv1~\cite{zhang2021fairmot}}& \revised{67.9}& \revised{68.1}& \revised{56.3}& \revised{32,571}& \revised{144,126}& \revised{4,293}\\
        \revised{FairMOTv2~\cite{zhang2021fairmot}}& \revised{73.7}& \revised{72.3}& \revised{\underline{59.3}}& \revised{27,507}& \revised{117,477}& \revised{3,303}\\
        \revised{QDTrack~\cite{pang2021quasi}}& \revised{68.7}& \revised{66.3}& \revised{53.9}& \revised{26,859}& \revised{146,643}& \revised{3,378}\\
        MO3TR (ours) & 63.9 & 60.5 & 49.9 & 23,358 & 177,684  & \underline{2{,}847}  \\
        MO3TR-PIQ (ours) & \textbf{77.6} & \underline{72.9} & \textbf{60.3} & \underline{21{,}045} & \underline{102{,}531}  & \underline{2{,}847}  \\ \bottomrule
    \end{tabular}
    }
    }
    \end{center}
    \vspace{-.7em}
\end{table}

\begin{table}
    \vspace{-0.5em}
    \begin{center}
    \caption{Results on the MOT20 benchmark~\cite{mot20} test set using \textbf{private} detections. \textbf{Bold} and \underline{underlined} numbers indicate best and second best result, respectively. Note that evaluation on the MOT20 benchmark does not differentiate between public and private detections (we use our private ones here). More detailed results of our approach are provided in the supplementary material.}
    \vspace{-.5em}
    \label{tab: MOT20}
    \scalebox{0.95}
    {
    {
    \setlength{\tabcolsep}{3pt}
    \begin{tabular}{@{}l *5c r r c@{}} 
    \toprule
           Method  & {MOTA$\uparrow$} & {IDF1$\uparrow$} & {HOTA$\uparrow$}   & {FP$\downarrow$}  & {FN$\downarrow$} & {IDs$\downarrow$}\\    \midrule  
        \midrule
        GSDT-V2 ~\cite{wang2021joint} &67.1  &67.5&53.6 &31,507  &135,395 &3,230  \\
        OUTrack ~\cite{liu2022online} &68.6  &\textbf{69.4}	& \underline{56.2} & 36,816  & \textbf{123{,}208} & {2{,}223} \\
        CrowdTrack ~\cite{stadler2021performance} & \underline{70.7}  & 68.2	& 55.0 & {21{,}928} & \underline{126{,}533}	 & 3,198 \\
        \revised{FairMOTv1~\cite{zhang2021fairmot}}& \revised{58.7}& \revised{63.7}& \revised{--}& \revised{--}& \revised{--}& \revised{6,013}\\
        \revised{FairMOTv2~\cite{zhang2021fairmot}}& \revised{61.8}& \revised{67.3}& \revised{54.6}& \revised{103,440}& \revised{88,901}& \revised{5,243}\\
        \revised{SOTMOT~\cite{zheng2021improving}}& \revised{{68.6}}& \revised{71.4}& \revised{--}& \revised{57,064}& \revised{101,154}& \revised{4,209}\\
        \revised{CorrTracker~\cite{wang2021multiple}}& \revised{{65.2}}& \revised{69.1}& \revised{--}& \revised{79,429}& \revised{95,855}& \revised{5,183}\\
        \revised{TrackformerV3~\cite{trackformer}}& \revised{{68.6}}& \revised{65.7}& \revised{--}& \revised{\underline{20,348}}& \revised{140,373}& \revised{\textbf{1,532}}\\
        MO3TR-PIQ (ours) & \textbf{72.3} & \underline{69.0} &\textbf{57.3}  & \textbf{12,738} & 128{,}439  & \underline{2,200}  
        
        \\ \bottomrule
    \end{tabular}
    }
    }
    \end{center}
    \vspace{-.7em}
\end{table}

\noindent\textbf{Public detections.} We evaluate MO3TR on the challenging MOT16 and MOT17 benchmark test datasets~\cite{mot17} using the provided public detections and report our results in Tables~\ref{tab:MOT16} and \ref{tab:MOT17}, respectively. Despite not using any heuristic track management to filter or post-process, we achieve competitive performance across both benchmarks. We outperform most competing methods regarding several metrics, achieving new state of the art results for MOTA, FN, MT and ML metrics on MOT16. On MOT17, we set new benchmarks regarding MOTA, HOTA and ML when we incorporate the public detections via the intersection over union (IoU) as proposed in~\cite{trackformer}, and regarding IDF1, HOTA and ML when using the center distance incorporation proposed in~\cite{zhou2020centertrack}.\par
As shown by its competitive IDF1 scores on both datasets, MO3TR is capable of identifying objects and maintaining their identities over long parts of the track, in many cases for more than 80\% of the objects' lifespans as evidenced by the comparably high MT and low ML results. We attribute this to MO3TR's access to the track history through the temporal Transformers. This information allows our method to jointly reason over existing tracks, possible initiation and the input data via our spatial Transformers, a capability that helps to learn discerning occlusion from termination, and thus to avoid false termination as is evidenced by the low FN and state of the art ML numbers achieved on both MOT datasets. These values further indicate that MO3TR learns to fill in gaps due missed detections or occlusions, which has additional influence on reducing FN and ML while increasing IDF1 and MT. Using its joint reasoning over the available information also helps MO3TR to reduce failed track initiations (FN) considerably while keeping incorrect track initiations (FPs) at a reasonably low levels. The combination of superior IDF1, low FN and reasonable FP allows MO3TR to reach new state of the art MOTA results on both MOT16~(Table~\ref{tab:MOT16}) and MOT17~(Table~\ref{tab:MOT17}) datasets.\par
Comparing our results specifically to the concurrently developed method Trackformer~\cite{trackformer}, we achieve comparable or better results across all metrics while using a significantly smaller backbone then both Trackformer versions (cf. Table~\ref{tab:MOT17}). Some of the improvements from TrackformerV1 towards its second version can be attributed to the switch to a multi-scale set detection concept~\cite{zhu2020deformable}, while we employ a single scale~\cite{carion2020detr} in our method (similar to TrackformerV1). It is to be noted that Trackformer~\cite{trackformer} exploits information only across two frames, while our MO3TR considers longer-term information. To make processing of our embedding sequences computationally tractable for any track length given constrained computational resources, we opted for a smaller backbone (ResNet50) for MO3TR. We expect our method to yield further improvements with the incorporation of bigger backbones and multi-scale detection concepts like~\cite{zhu2020deformable}, however at higher computational costs.\par
\vspace{4pt}
\noindent \textbf{Private detections.} To allow for broader comparability of our approach to other recent methods, we also evaluate MO3TR on the MOT17 benchmark using our method's own \textit{private} detections, achieving competitive results \ignore{compared to other current state of the art Transformer-based methods }(cf. Table~\ref{tab: MOT17Private}). \par
\vspace{4pt}
\noindent \textbf{MO3TR with predicted initiation queries.} \revised{As introduced in Section~\ref{sec:extension_zq}, we additionally investigate how modifying the encoder of our approach as described (multi-scale backbone, additional auxiliary detection loss) and dynamically predicting the initiation queries~$\mathcal{Z}_Q$ conditioned on each input image instead of learning a fixed set of queries will influence the performance of our approach. We refer to this method as MO3TR-PIQ (\underline{P}redicted \underline{I}nitiation \underline{Q}ueries).} \par
\revised{The results achieved on the MOT17 benchmark (Table~\ref{tab: MOT17Private}) show significant gains in performance across all metrics, including 13.7\% on MOTA  and 10.4\% on HOTA compared to our standard method -- setting new state of the art results across serveral metrics (MOTA of 77.6\%).} Evaluations conducted on the recently proposed and rather complex MOT20 benchmark underpin the performance capabilities by outperforming competing methods on several metrics (Table~\ref{tab: MOT20}\revised{, please refer to the supplementary material for further results using public detections}). We mainly attribute this improvement on the benchmarks to the fact that set-based detection methods like DETR~\cite{carion2020detr} have just very recently been introduced to the community, and are still often significantly outperformed by non-set based ones like~\cite{ge2021yolox}. While MO3TR is entirely set-based and built upon the detection concept of DETR, MO3TR-PIQ is incorporating the more powerful multi-scale detection setup of~\cite{ge2021yolox} into our tracking method -- demonstrating potential for future combinations with other detection methods.\par
\vspace{4pt}
\noindent \textbf{Inference speed vs.$\:$accuracy.}
\revisedminor{We provide details regarding the inference speed-accuracy comparison in Table~\ref{tab:MOT17_speed}. While center-based methods like CenterTrack~\cite{zhou2020centertrack} and FairMOT~\cite{zhang2021fairmot} seem to generally provide faster inference (possibly due to their point-based interpretation of the tracking problem), both our models show inference times that are comparable to the other two Transformer-based methods. While being comparable in performance (MOTA and IDF1), our base MO3TR method is notably faster than TrackformerV2~\cite{trackformer} ($11.9$ FPS vs. $7.4$ FPS). Our extended MO3TR-PIQ approach is slightly slower than the base version and is positioned directly between TrackformerV2/V3~\cite{trackformer} and TransTrack~\cite{sun2020transtrack} in terms of inference speed while outperforming both in terms of MOTA and IDF1.} 
\begin{table}
    \begin{center}
    \caption{\revised{Comparing inference speed and performance on the MOT17 benchmark~\cite{mot17} test set using private detections.}}
    \label{tab:MOT17_speed}
    \vspace{-.5em}
    \scalebox{0.95}
    {
    \setlength{\tabcolsep}{3pt}
    \begin{tabular}{@{}l *2c r@{}} 
    \toprule
           Method  & {MOTA$\uparrow$} & {IDF1$\uparrow$} & {FPS$\uparrow$}\\    \midrule  
        \revised{FairMOTv2}~\cite{zhang2021fairmot} & $73.7$ & $72.3$ & $25.9$\\
        \revised{CenterTrack}~\cite{zhou2020centertrack} & $67.8$ & $64.7$ & $17.7$\\
        \revisedminor{TrackformerV2}~\cite{trackformer} & $65.0$ & $63.9$ & $7.4$ \\
        \revisedminor{TrackformerV3}~\cite{trackformer} & $74.1$ & $68.0$ & $7.4$ \\
        \revised{TransTrack}~\cite{sun2020transtrack} & $74.5$ & $63.9$ & $10.0$ \\
        \revised{MO3TR (ours)} & $63.9$ & $60.5$ & $11.9$\\
        \revised{MO3TR-PIQ (ours)} & $77.6$ & $72.9$ & $8.8$ \\
        \bottomrule
    \end{tabular}
    }
    \end{center}
\end{table}

\subsection{Ablation studies}
\label{sec:ablations}
\begin{table}
    \begin{center}
    \vspace{-.8em}
    \caption{Effect of varying lengths of track history~$\mathcal{T}^m_{t-1}$ considered in the temporal Transformers during evaluation. Ablation studies were performed using our validation set combining indoor and outdoor sequences of MOT20~\cite{mot20} and 2DMOT15~\cite{2dmot15} to cover a representative range of different pedestrian density levels.}
    \label{tab: buffer}
    \vspace{-.5em}
    \scalebox{0.95}
    {
    \setlength{\tabcolsep}{7pt}
    \begin{tabular}{@{} l *6c@{}}
    \toprule
        {len($\mathcal{T}^m_{t-1}$)} & {MOTA$\uparrow$} & {IDF1$\uparrow$} & {MT$\uparrow$} & {ML$\downarrow$} & {FP$\downarrow$} & {FN$\downarrow$} \\              
        \midrule
            ~~1  & 55.4 & 48.4 & 114 & 19 & 4,700 & 12,898  \\
            10 & 56.8 & 49.0 & 115 & \textbf{18} & 4,245 & 12,805  \\
            20 & 57.8 & 50.1 & 115 & 19 & 3,826 & 12,787  \\
            30 & \textbf{58.9} & \textbf{50.6} & 114 & 20 & \textbf{3{,}471} & \textbf{12{,}692}  \\
        \bottomrule
    \end{tabular}
    }
    \end{center}
     \vspace{-.8em}
\end{table}

\begin{table}
    \begin{center}
    \caption{Effect of different training (two frames vs. 30) and augmentation strategies: False Negatives (FN), False Positives (FP), Right-Aligned insertion (RA). Ablation studies were performed using our validation set combining indoor and outdoor sequences of MOT20~\cite{mot20} and 2DMOT15~\cite{2dmot15} to cover a representative range of different pedestrian density levels.}
    \label{tab: training}
    \vspace{-.5em}
    \scalebox{0.95}
    {
    \setlength{\tabcolsep}{6pt}
    \begin{tabular}{@{} l *2c *2r@{}}
    \toprule
        Training Strategies & {MOTA$\uparrow$} & {IDF1$\uparrow$} & {FP$\downarrow$} & {FN$\downarrow$} \\              
        \midrule
        Naive (Two Frames) & 12.2 & 22.1 &  7,905 & 26,848  \\
        FN (Two Frames)  & 14.6 & 42.0 & 22,609 & \textbf{11{,}671}  \\
        FN+RA (Two Frames) & 28.4 & 42.5  & 16,749 & 11,940  \\
        FN+RA+FP (Two Frames) & 55.4 & 48.4 & 3,927 & 17,912  \\
        FN                  & 21.9 & 42.5 & 19,353 & 11,693  \\
        FN+RA               & 39.2 & 48.1 & 12,265 & 12,002  \\
        FN+RA+FP            &\textbf{58.9} &\textbf{50.6} &\textbf{3{,}471} & 12,692  \\
        \bottomrule
    \end{tabular}
    }
    \end{center}
\end{table}

\begin{figure*}[ht]
 \begin{center}
  \scalebox{1.0}{
  \put(-1.5,117){\rotatebox{90}{\cite{tracktor}}}
  \put(-1.5,250){\rotatebox{90}{\cite{tracktor}}}
  \includegraphics[width=\linewidth]{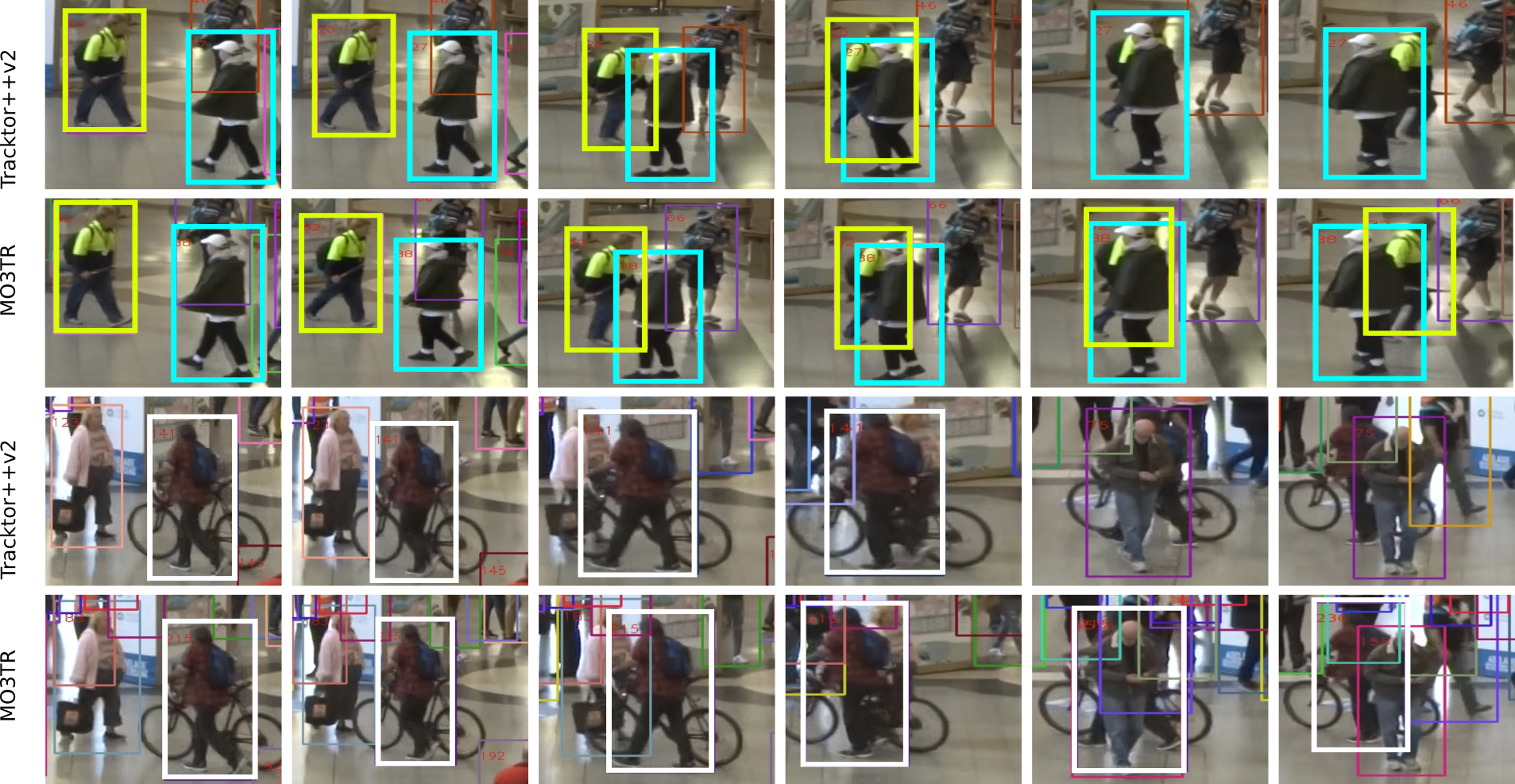}
}
\end{center}
  \caption{Qualitative results of two challenging occlusion scenarios in the validation set. Objects of focus are highlighted with slightly thicker bounding boxes. Unlike Tracktor++v2~\cite{tracktor}, our proposed MO3TR is capable of retaining the identity and keeps track even if the object is severely occluded.}
\label{fig:qual_res}
\end{figure*}

In this section, we evaluate different components of MO3TR on our validation set using private detections and show the individual contributions of the key components and strategies to facilitate learning. \par
\vspace{4pt}
\noindent \textbf{Effect of track history length.} The length of the track history describes the maximum number of embeddings from all the previous time steps of a certain identified object that our temporal Transformer has access to. To avoid overfitting to any particular history length that might be dominant in the dataset but not actually represent the most useful source of information, we specifically train our model with input track histories of varying and randomly chosen lengths. It is important to note that if the maximum track history length is set to one, the method practically degenerates to a two-frame based joint detection and tracking method such as Trackformer~\cite{trackformer}. Our results reported in Table~\ref{tab: buffer} however show that incorporating longer-term information is crucial to improve end-to-end tracking. Both MOTA and IDF1 can be consistently improved while FP can be reduced when longer term history, \ie, information from previous frames, is taken into account. This trend is also clearly visible throughout evaluation of our training strategies presented in Table~\ref{tab: training}, further discussed in the following. \par
\vspace{4pt}
\noindent\textbf{Training strategies.} MOT datasets are highly imbalanced when it comes to the occurrence of initialization and termination examples compared to normal propagation, making it nearly impossible for models to naturally learn initiation of new or termination of no longer existing tracks when trained in a naive way. As presented in Table~\ref{tab: training}, naive training without any augmentation shows almost double the number of false negatives~(FN) compared to augmented approaches, basically failing to initiate tracks properly. Augmenting with FN as discussed in~\ref{sec:training} shows significant improvements for both two-frame and longer-term methods. Additionally right-aligning the track history helps generally to stabilize training and greatly reduces false positives. At last, augmenting with false positives is most challenging to implement but crucial. As the results demonstrate, it significantly reduces false positives by helping the network to properly learn the terminating of tracks. \par
\begin{figure}[t]
    \begin{center}
    {
        \includegraphics[width=1\linewidth]{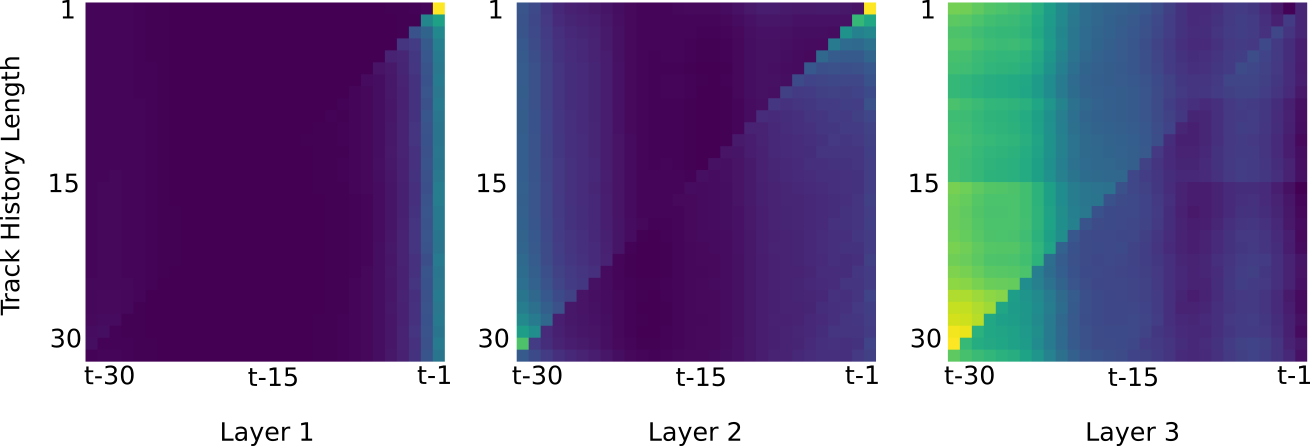}
    }
    \end{center}
    \vspace{-.9em}
   \caption{Temporal attention maps averaged over 100 randomly selected objects from the MOT20 dataset~\cite{mot20}. The vertical axis represents the maximum track history length, the horizontal axis the different embedding positions in the history. The displayed attention related the current query at time $t$ to all the previous embeddings. Every row sums up to 1.}
   \vspace{-1.0em}
   \label{fig:temporal attention}
\end{figure}
\vspace{4pt}
\noindent\textbf{Analysing temporal attention.}
To provide some insight into the complex and highly non-linear working principle of our temporal Transformers, we visualize the attention weights over the temporal track history for different track history lengths averaged for 100 randomly picked objects in our validation set  (Fig.~\ref{fig:temporal attention}). Results for the first layer clearly depict most attention being payed to multiple of its more recent frames, decreasing with increasing frame distance. The second and third layers are harder to interpret due to the increasing non-linearity, and the model starts to increasingly cast attention over more distant frames. It is important to notice that even if an embedding is not available at time $t-k$, the model can still choose to pay attention to that slot and use the non-existence for reasoning. 
\section{Conclusion}
We presented MO3TR, a truly end-to-end multi-object tracking framework that uses temporal Transformers to encode the history of objects while employing spatial Transformers to encode the interaction between objects and the input data, allowing it to handle occlusions, track termination and initiation. Demonstrating the advantages of long term temporal learning, we set new state of the art results regarding multiple metrics on the popular MOT16, MOT17 and MOT20 benchmarks.

\noindent\textbf{Future work.}
\revisedminor{
It is known to the tracking community that most popular datasets and metrics overemphasize detection performance (see supplementary).
This problem has recently been tackled by the introduction of the new $\mathrm{HOTA}$ metric in~\cite{luiten2021hota}, which is designed to better balance the contributions. While developing novel metrics is one valuable approach, we propose to potentially break with this tradition in the future and instead reconsider the data we use as basis to evaluate tracking. 
Creating data with significantly easier detection (medium/large object sizes) but much harder tracking tasks (complex movements including frequent occlusions and interactions) could help to better decouple the contributions of tracking approaches and motivate the community to focus on more diverse aspects of tracking beyond detection as well as broaden the access due to the reduced computational complexity.}

\noindent\textbf{Acknowledgements.} The authors gratefully acknowledge the support of the ARC through projects CE140100016, FL130100102 and DP200102427.

\ifCLASSOPTIONcaptionsoff
  \newpage
\fi



\bibliographystyle{IEEEtran}
\bibliography{IEEEabrv, main.bib}

\vspace{-3em}
\begin{IEEEbiography}
    [{\includegraphics[width=1in,height=1.25in,clip,keepaspectratio, trim={0 2mm 0 10mm}]{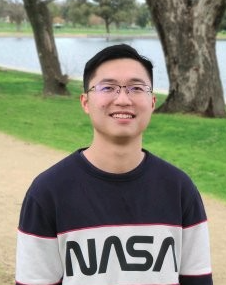}}]{Alan Tianyu Zhu}
is a computer vision PhD candidate at Monash University since 2018. He has received Bachelor of Engineering and Science from Monash University with first class Honours in 2017. He is also a machine learning research associate at Faculty of IT, Monash University on learning with less labels projects. His current research focus is attention mechanisms, object detection and multi object tracking. 
\end{IEEEbiography}

\vspace{-2.5em}
\begin{IEEEbiography}
    [{\includegraphics[width=1in,height=1.25in,clip,keepaspectratio]{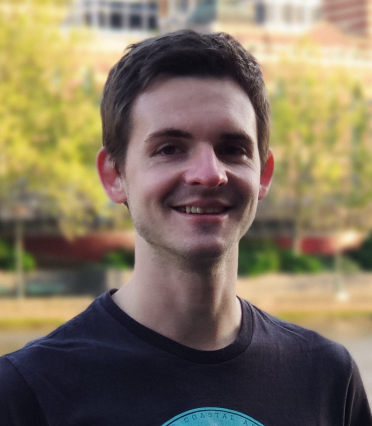}}]{Markus Hiller} is currently working towards his PhD at the School of Computing and Information Systems at the University of Melbourne, Australia. Prior to this, he has worked for 2.5 years on a research project studying environment perception and modelling for mobile robots after having received a Bachelor of Science and Master of Science from FAU, Germany in 2015 and 2017, respectively. His current research interests include representation learning, learning from limited data and under distribution shifts, geometry, as well as various applications in computer vision. 
\end{IEEEbiography}

\begin{IEEEbiography}
    [{\includegraphics[width=1in,height=1.25in,clip,keepaspectratio]{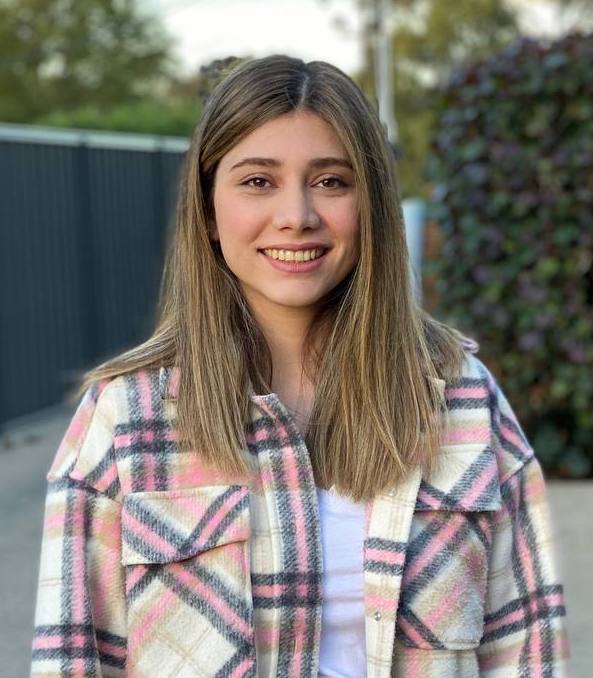}}]{Mahsa Ehsanpour} is a PhD candidate in Computer Science at The University of Adelaide and the Australian Institute for Machine Learning (AIML) since 2018. Her research interest lies at the intersection of deep learning applied to computer vision applications with a specific focus on understanding humans in motion within video sequences. More specifically, her research encompasses fine-grained human understanding such as human pose, body surface and trajectory prediction as well as coarse-grained human understanding such as comprehending human actions, interactions with other individuals (social groups) and social activities.
\end{IEEEbiography}

\vspace{-3em}
\begin{IEEEbiography}
    [{\includegraphics[width=1in,height=1.25in,clip,keepaspectratio, trim={0 2cm 0 0}]{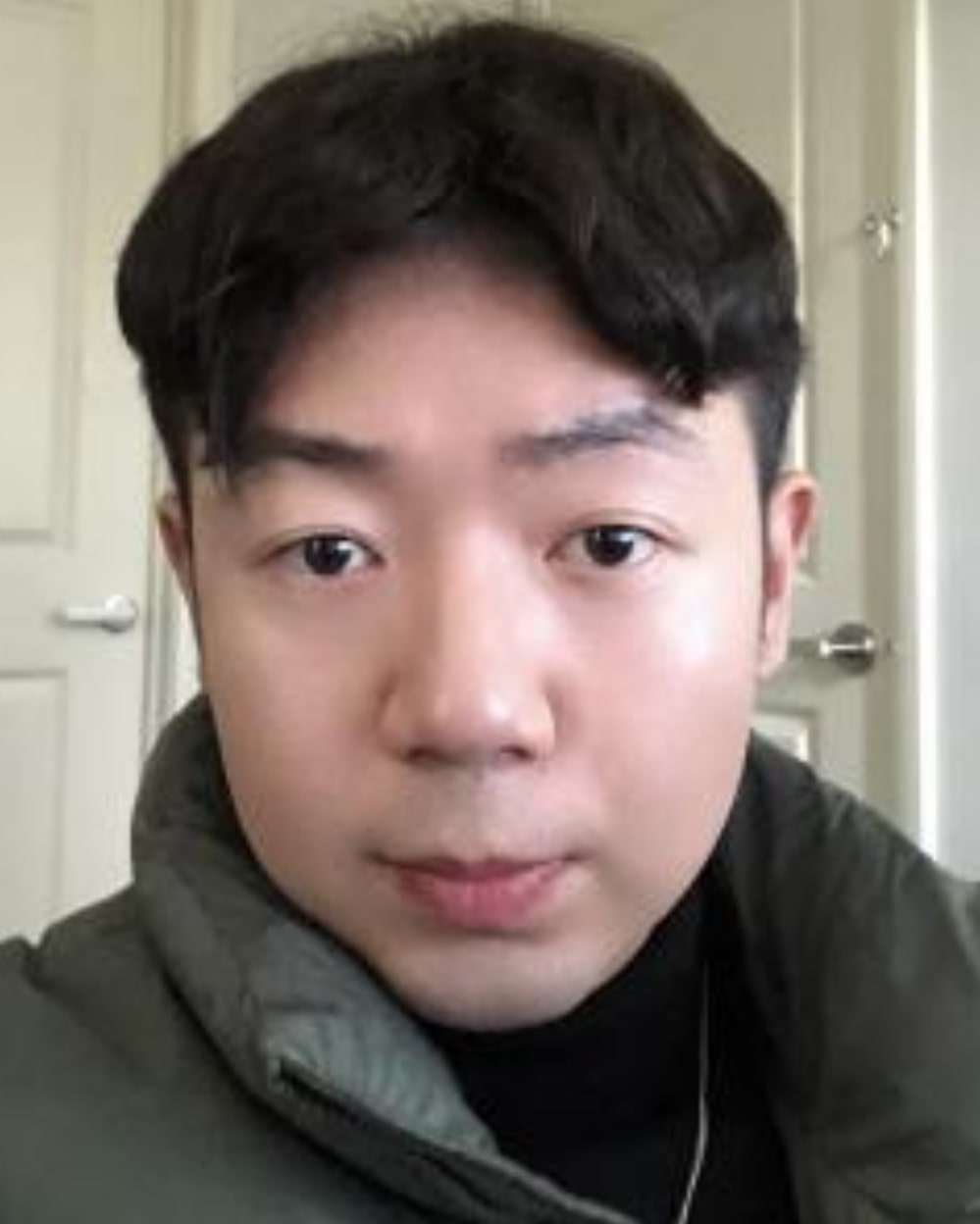}}]{Rongkai Ma} is a PhD candidate at Monash University and Australian Research Council Centre of Excellence for Robotic Vision since 2019. Before he started his PhD, he obtained the Bachelor's degree in Electrical and Computer Systems Engineering (ECSE) at Monash University. His research interest focuses on learning adaptively with few observations for computer vision applications.
\end{IEEEbiography} 

\vspace{-3em}
\begin{IEEEbiography}
    [{\includegraphics[width=1in,height=1.25in,clip,keepaspectratio]{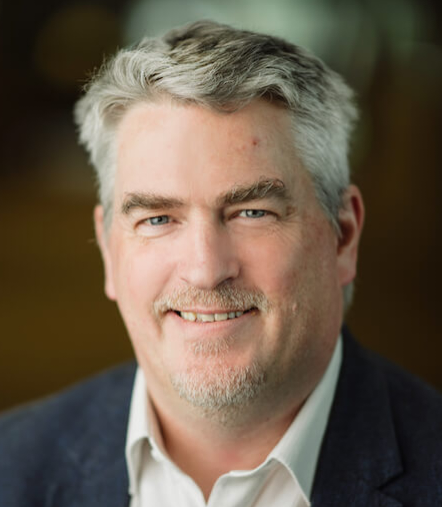}}]{Tom Drummond} is the Melbourne Connect Chair of Digital Innovation for Society at the University of Melbourne.  Between 2010 and 2021, he was a Professor of Engineering and from 2016, the Head of Department of Electrical and Computer Systems Engineering at Monash University. From 2001 to 2010, he was a Lecturer in Engineering at the University of Cambridge. His research interests include Robotic Vision, Augmented Reality, Machine Learning and Efficient Algorithms. He has published more than 150 peer reviewed papers and serves on the program committees of numerous national and international conferences and was the General Chair of the International Symposium of Mixed and Augmented Reality ISMAR in 2008.
\end{IEEEbiography}

\vspace{-3em}
\begin{IEEEbiography}
    [{\includegraphics[width=1in,height=1.25in,clip,keepaspectratio]{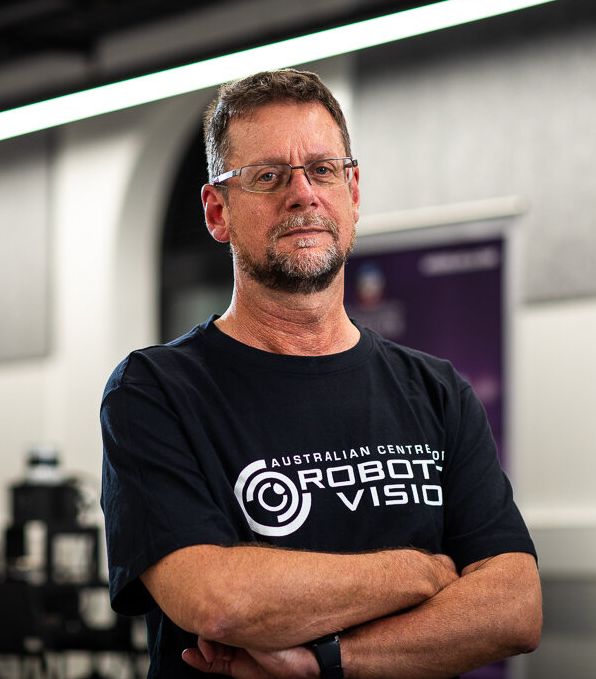}}]{Ian Reid} is the Head of the School of Computer Science at the University of Adelaide. He is a Fellow of Australian Academy of Science and Fellow of the Academy of Technological Sciences and Engineering, and held an ARC Australian Laureate Fellowship 2013-18. Between 2000 and 2012 he was a Professor of Engineering Science at the University of Oxford. His research interests include robotic vision, SLAM, visual scene understanding and human motion analysis. He serves on the program committees of various national and international conferences, and was an Area Editor for T-PAMI, 2010-2017.
\end{IEEEbiography}

\vspace{-3em}
\begin{IEEEbiography}
    [{\includegraphics[width=1in,height=1.25in,clip,keepaspectratio]{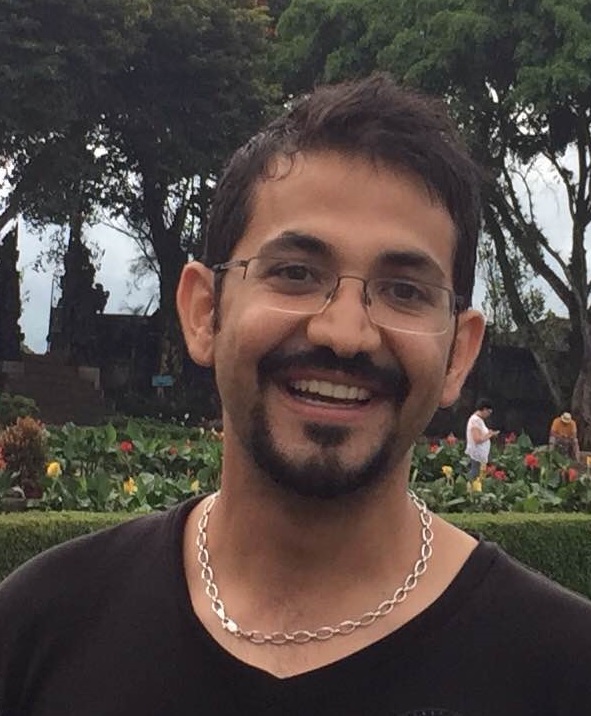}}]{Hamid Rezatofighi}
 is a lecturer at the Faculty of Information Technology, Monash University, Australia. Before that, he was an Endeavour Research Fellow at the Stanford Vision Lab (SVL), Stanford University, and a Senior Research Fellow at the Australian Institute for Machine Learning (AIML), the University of Adelaide. He received his PhD from the Australian National University in 2015.  He has published over 60 top tier papers in computer vision, AI and machine learning, robotics, and has been involved as primary investigator in several successful grant fundings (totally over \$11.0M), including two DARPA grants and one ARC discovery grant. He served as the publication chair in ACCV 2018 and as the area chair in  WACV 2021, CVPR 2020 and CVPR 2022-2023. His research interests include computer vision, machine learning and robotic vision, \emph{esp.} the visual perception problems that are required for an autonomous robot to navigate or interact in a human crowded environment.  
\end{IEEEbiography}




\setcounter{table}{0} \renewcommand{\thetable}{\Alph{section}\arabic{table}}
\setcounter{figure}{0} \renewcommand{\thefigure}{\Alph{section}\arabic{figure}}
\setcounter{section}{0}\renewcommand\thesection{\Alph{section}}

\section{Further details on training MO3TR}
\revisedminor{In this section we provide additional details regarding the training datasets, two different training approaches and the training complexity for our method.}
\subsection{Training datasets and training procedure}
\revisedminor{We provide some further details regarding the multi-stage approach described in the main paper in the following. We additionally analysed a single-stage training strategy where training of detection and tracking is performed jointly at once and report the results in Table~\ref{tab:multisinglestage}.}\par
\vspace{4pt}
\noindent\revisedminor{\textbf{Multi-stage training.} The exact training procedure slightly differs between our base MO3TR and our extended MO3TR-PIQ models. While both backbones are initialised with weights pretrained on COCO~\cite{lin2014coco}, the MO3TR encoder and spatial Transformers are first trained on a composition of the CrowdHuman~\cite{crowdhuman}, ETH~\cite{ethdataset} and CUHK-SYSU~\cite{xiao2017joint} datasets on a detection task (stage 1) and then trained together with the engaged temporal Transformer on the respective MOT tracking dataset (stage 2). As detailed in the main paper, this is easily achievable due to the comparably small size of our overall model even with rather limited computational resources. However, due to the increased size of the multi-scale backbone and encoder used in our extended MO3TR-PIQ model, we attempted to limit the computational complexity and training time to better suit limited-resource settings while still capturing the essential information to learn detection and tracking. We thus limited our dataset used during the first stage of training (detection task) to only contain the CrowdHuman dataset commonly used for detection tasks and complemented this with the rather small but diverse task-specific MOT dataset (MOT17~\cite{mot17} or MOT20~\cite{mot20}). For stage 2, we follow the identical approach as before and train on the respective task-specific MOT dataset but freeze the backbone to reduce the computational complexity. Since using the reduced dataset combined with the frozen backbone already yielded promising results and is computationally feasible for setups with limited resources, we opted to keep this data-reduced training setup throughout our MO3TR-PIQ experiments.}\par
\vspace{4pt}
\noindent\revisedminor{\textbf{Single-stage training.} To gain insights into how our method performs for a different training setup that requires increased computational resources, we adopted a single-stage training strategy to train the entire MO3TR-PIQ model in one stage end-to-end with unfrozen backbone (initialised with COCO-pretrained weights). In detail, each batch used in training is composed of two sub-batches: one containing samples of the detection and the other of the tracking task. We compute the individual losses and simply add them to form the total loss. The results reported in Table~\ref{tab:multisinglestage} demonstrate that single-stage training with unfrozen backbone is feasible and can further improve the overall performance, however comes at a higher computational cost (in our experiments 4x Quadro RTX6000 with 24GB each).}
\begin{table*}
    \begin{center}
    \caption{Comparison between multi-stage trained MO3TR-PIQ (MST) and single-stage trained MO3TR-PIQ (SST) on MOT17 private.}
    \label{tab:multisinglestage}

    \begin{tabular}{lcccccc} 
        \toprule
        Method  & {MOTA$\uparrow$}  & {IDF1$\uparrow$} & {HOTA$\uparrow$}  & {FP$\downarrow$}  & {FN$\downarrow$} & {IDs$\downarrow$}\\    \midrule 
        MO3TR-PIQ (MST) & 77.6& \textbf{72.9} & 60.3 & \textbf{21,045} & 102,531 & 2,847\\
        MO3TR-PIQ (SST) & \textbf{78.6} & 72.4 & \textbf{60.5} & 24,873 & \textbf{93,174} & \textbf{2,808}\\
        \bottomrule
    \end{tabular}

    \end{center}
\end{table*}

\subsection{Model and training complexity}
\revisedminor{Similar to other tracking works, our method can be divided into the backbone that extracts and encodes the visual information from the images and the actual tracking module. The number of parameters used within our actual tracking part, \ie Temporal and Spatial Transformers, is relatively small compared to the common ResNet50 backbone ($\sim$25M parameters) plus Encoder~($\sim$7.9M parameters) -- meaning that our actual tracking module doesn't add much computational overhead due to its comparably small size. In detail, our MO3TR method's Temporal Transformer is an 8-head 3-layer architecture that adds $\sim$1.6M parameters, while our Spatial Transformer module consists of 6 pairs of self-attention and cross-attention modules contributing a total of $\sim$9.5M parameters. This results in $\sim$33M parameters for the backbone, and $\sim$11M for our tracking part.\ignore{ -- a parameter count that is in total similar to the number of parameters for the backbone used in other works like Tracktor~\cite{tracktor} (ResNet101 $\sim$44.5M).}\\
As detailed in Section~4.1 in the paragraph "Computational complexity for training" in the main paper, we train the entire MO3TR model using 4~GTX 1080ti with 11GB each, \ie totalling 44GB of maximum available memory -- which is significantly smaller than most other recently published works\ignore{ (\eg~Trackformer~\cite{trackformer} using "\textit{7 × 32GB GPUs}" for 2~days to train the detection module)}.
We trained our MO3TR model with a batch size of 32 images for 300 epochs during the first stage (pedestrian detection task), using up the available 44GB spread across our 4 GPUs for slightly less than 6 days. For the second stage of our procedure (with engaged temporal Transformer), we trained the model end-to-end on a single GPU for 100 epochs with a batch size of 4, requiring a total of 11GB. We would like to note that training times are not necessarily representative of complexity, since identical code executed on two different types of hardware (\eg V100 vs. newer A100 GPUs) can lead to significantly different times.}

\section{MO3TR -- Design choices and insights}
\revisedminor{
In this section, we share some additional insights into the inner workings of our method and our design choices.}
\subsection{Transformers vs. RNNs for temporal information}
\revisedminor{There has been a rich history of works using temporal information to tackle tracking problems, and we would like to re-iterate the we are not claiming to be the first to do so. The novel use of Transformers to incorporate this information as employed in our work however does offer significant benefits over RNNs: 1) Transformers in general overcome the fundamental constraint of sequential computation that is inherent to recurrent methods (like RNNs, LSTMs, GRUs)~\cite{attentionisallyouneed}. Our temporal Transformers can process the available object embeddings of all time steps in parallel without the need to unroll or sequentially process the information. Having access to the entire sequence at once allows our method to `skip' or exclude unhelpful embeddings of certain time steps in a natural way -- an ability that is much harder to achieve with the sequential nature of recurrent methods. 2) Training our temporal Transformer module proved very stable and efficient (also given their small size as detailed in Section~4.1), whereas training RNNs is often described as rather difficult~\cite{pascanu2013difficultyrnn} and inefficient due to the involved unrolling during training. This holds true especially for longer sequences. We thus believe that leveraging the capability of Transformer architectures to process the available temporal information in parallel will prove advantageous especially if longer sequences are involved, and consider it a better choice to process temporal information going forwards.}
\subsection{Leveraging longer sequences beyond two frames}
\revisedminor{The introduction of using multiple frames during the training process poses several advantages over the commonly employed two-frame approach. Most recent tracking works attempt to learn track initiation and termination by sampling two frames out of a short sequence. This is however only a sparse and often poor approximation of the actual complex tracking scenario and thus frequently requires significant post-processing or dataset-specific tuning of the augmentation process (\eg adjusting the false positive rate) due to the gap between the scenarios encountered during training and at inference time. In contrast to this, we attempt to take a step towards closing this gap and use our model to infer the results for up to 30 consecutive frames during training. This naturally provides scenarios where occlusions commonly occur within the training sequence, something that simply cannot be resolved in only two frames. We then employ the same algorithm underlying the CLEAR metric~\cite{bernardin2008evaluating} to match our predicted tracklets and ground truth annotations. We believe that this procedure better replicates the scenarios that will likely be encountered during inference already during the training process, allowing for complex scenarios to naturally unfold and for errors to accumulate over multiple frames -- thus helping to learn the characteristics of the tracking problem from the data itself instead of dataset-specific augmentation parameters. We additionally complement this data-driven strategy by our proposed augmentations.}\par
\begin{figure}[t]
    \begin{center}
    {
        \includegraphics[width=1\linewidth]{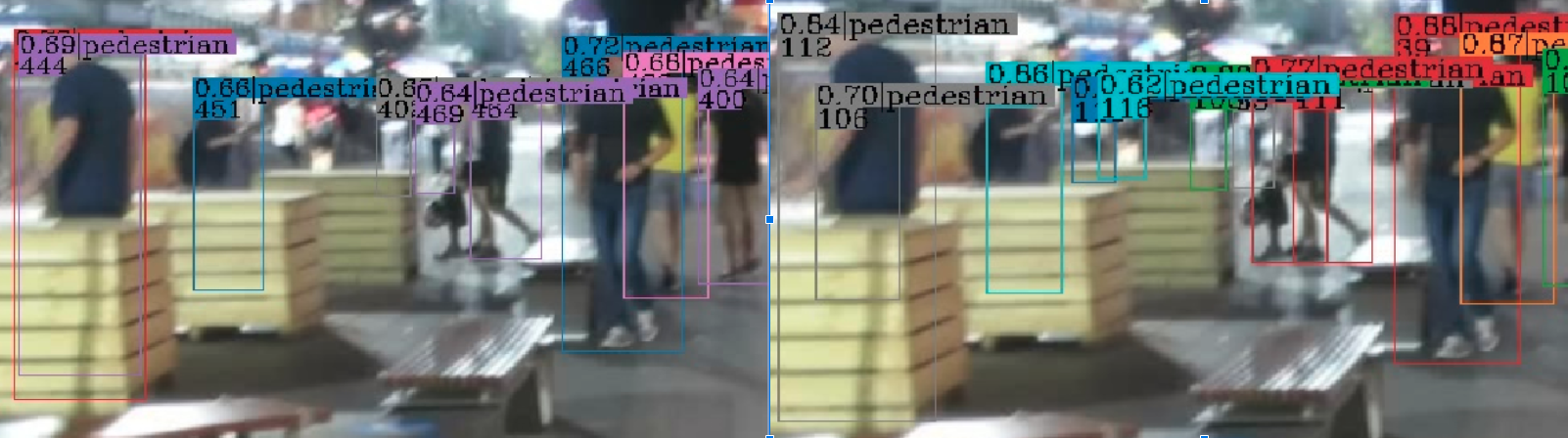}
    }
    \end{center}
  \caption{Comparing the tracking outcomes for our model trained with a common naive two-frame (left) and our proposed multi-frame approach (right). Both models have tracked objects for 300 frames with the results of the final frame depicted. The two-frame model has a maximum ID number of 469, indicating high occurrence of false positives. The multi-frame method shows significantly lower IDs, indicating its ability to successfully mitigate this problem. (Note that box colors are not indicative of individual object identities in this figure.)}
  \label{fig:dup}
\end{figure}

\revisedminor{
One important example where the multi-frame training strategy comes into play is learning track termination in complex multi-pedestrian scenarios: Assume there are 50 objects present in frame~1 and 49 objects in frame~2, \ie one person's track ought to be terminated. Considering that the loss is computed across all objects, the contribution of a `correct termination' or penalty of `incorrect tracking' is rather small compared to the successful tracking of the other objects. This data imbalance is a common problem that leads the network to take the `easier' path of simply keeping all objects, which requires extensive post-processing or dataset specific adaptation of the augmentation process to create such balance. Figure~\ref{fig:dup} depicts our model trained via a 2-frame approach on the left, showing very high ID numbers with a maximum of 469 which indicate a great number of false positives (missed terminations). Even though not visible (due to perfect overlap of the boxes), the model actually keeps almost all objects alive and simply does not learn termination, yielding new detections with new IDs for many frames. In contrast, the right part of Figure~\ref{fig:dup} shows the identical sequence evaluated with our method trained using our multi-frame approach, with IDs significantly smaller due to correctly learnt object termination, since small errors that are not instantly penalised can accumulate over multiple frames to counteract the data imbalance problem. For brief discussion and quantitative results, please also see Section~3.5 and Table~7 of the main paper. }
\subsection{Avoiding the initiation of duplicate tracks}
\revisedminor{While the previous section demonstrates how track initiation, termination and occlusion handling are learnt in a more general way by using longer frame sequences, we provide further details on the architectural components of our method that enable it to learn avoiding the initiation of duplicates in this section.
As described in Section~3.3, we concatenate the set of embeddings~$\hat{\mathcal{Z}}_t$ predicted by the Temporal Transformer with the set of initiation queries~$\mathcal{Z}_Q$ and provide them as input~$\Bar{Z}_t$ to the Spatial Transformers (Fig.~2 and Eq.$\,$(3) in the main paper). The way how considering these sets jointly helps to avoid duplicates can be understood as follows: As detailed in the previous section (Point~2.1), the role of the initiation queries can be interpreted as learning to propose where to look for new, yet untracked objects. At the same time, the predicted embeddings~$\hat{\mathcal{Z}}_t$ provide information where and how already tracked objects can be located. These two pieces of initially disjoint information are naturally expected to frequently overlap, since tracked individuals will move into regions where new objects are expected to appear. Resolving this challenge and thus avoiding the creation of duplicates requires us to go slightly deeper into the way how our Transformer architecture processes the information: Initiation queries and embeddings predicted by the Temporal Transformer are both passed as input~$\Bar{\mathcal{Z}}_t$ to the Spatial Transformer, and thus create individual sets of keys, queries and values (one \{key, query, value\} triplet for each embedding, \cf Eqs.~(3) and~(4) as well as \cite{carion2020detr,attentionisallyouneed} for more detailed information). As described in Section~3.3 of our main paper, we perform self-attention over all embeddings in~~$\Bar{\mathcal{Z}}_t$, followed by cross-attention between the refined embeddings and the encoded image (both repeated for several times) and then predict the final set of bounding boxes. This process creates a direct relationship between the individual embeddings within~$\Bar{\mathcal{Z}}_t$ (explicitly modelling all pairwise interactions), as well as between the embeddings in~$\Bar{\mathcal{Z}}_t$ and the bounding boxes predicted for the input image. Since only one bounding box is assigned to any object as `correct' according to the provided ground truth and since our training method always prefers (and enforces) assignment of previously tracked objects over newly initialized ones, the parameters within the sequential refinement procedure throughout the Spatial Transformer can learn to either `reduce' the importance of initiation queries if well-matching/similar track-queries are already present in the sequence or to combine both information sources into one refined output embedding. This allows our method to naturally learn to avoid duplicates directly from the training data.}
\subsection{Termination of tracks}
\revisedminor{The termination of tracked objects within our approach comprises two different aspects: the learning-based information processing and the practical resource-considering implementation. Having access to the track history via the embeddings within the Temporal Transformer combined with the visual input information of the images provided by the encoder, our model can learn to combine motion information with visual appearance of tracked objects. Having all areas of the image explicitly encoded through our use of positional embeddings within the Spatial Transformer component (see Section~4.1) further enables our approach to naturally determine areas in the image where objects have a higher likelihood to `disappear' (\eg around the image boundaries). All this information is jointly considered when our model predicts the final output embeddings set~$\mathcal{Z}_t$, thus implicitly being aware of likely terminations. In practice and as further detailed in Section~4.1, we train our model with a maximum track history length of 30 frames, meaning that we can expect our model to be able to re-identify objects that might have been occluded for at most this number of frames. If an object has not been encountered after this time frame, we can assume the track has terminated. In practice, this means we keep tracklets for a maximum duration of 30 frames after they have been last assigned an output embedding~$z^m$ (see Fig.$\,$2 in the manuscript), and then remove them from our set of active tracks. It is to be noted that this 30 frame cut-off is however only a design choice to limit the required computational resources and does not affect the capability of our method to track and re-identify objects across longer time spans if provided with appropriate resources.}
\subsection{Potential alternatives to novel metrics}
\revisedminor{
It is known to the tracking community that most popular datasets and metrics~(\eg~\cite{mot17}) overemphasize detection performance. The commonly used $\mathrm{MOTA}$ is for example measured as}
\begin{equation}
    \mathrm{MOTA} = 1 - \frac{\sum_t \mathrm{FN}_t + \mathrm{FP}_t + \mathrm{IDSW}_t}{\sum_t \mathrm{GT}_t},
\end{equation}
\revisedminor{where $\mathrm{IDSW}$ represents identity switches, $\mathrm{FN}$ the false-negatives, $\mathrm{FP}$ false-positives and $\mathrm{GT}$ the number of objects present at time step $t$. A typical state of the art tracker for MOT17 will have scores for $\mathrm{FP}$ around 20k, $\mathrm{FN}$ around 150k and $\mathrm{IDSW}$ around 3k. Thus, the important tracking metric $\mathrm{IDSW}$ only contributes roughly 2\% and is thus highly dominated by the detection performance ($\mathrm{FN, FP}$). One reason for the imbalance is caused by most objects being either static or not engaged in any form of occlusion for most of the frames, leading to comparably small numbers of ID switches even for mediocre tracking. The detection, however, often poses a significant challenge in many MOT datasets with objects being of various sizes and appearances. Hence, many works focused on improving the $\mathrm{MOTA}$ through better detection results by using deeper and increasingly complex neural networks (along with increased computational complexity). While the public detection challenge made a good attempt to alleviate this problem by providing detections, this approach is no longer suited for modern joint detection and tracking frameworks.} \par
\revisedminor{This problem has recently been tackled by the introduction of the new $\mathrm{HOTA}$ metric in~\cite{luiten2021hota}, which is designed to better balance the subcomponents. While developing novel metrics is one valuable approach, we propose to potentially break with this and instead reconsider the data we use as basis for evaluating tracking quality.
One straight forward way we see that could significantly improve transparent evaluation of actual tracking performance in the future would be to create a dataset consisting of easier detection but much harder tracking tasks to decouple the contributions towards detection and tracking. Captured data should contain reasonable numbers of medium to large-sized objects that are easy to detect but engage in more complex interactions and movements (including entering and exiting the scene), thus naturally creating plenty of occlusions, direction changes and initiations/terminations. This would additionally alleviate data imbalance problems faced by many modern learnable tracking frameworks. We believe that tackling such datasets could motivate the community to focus on more diverse aspects of tracking beyond detection, and broaden the access due to the reduced computational complexity that is required.
}
\begin{table*}[h!]
    \begin{center}
    \scalebox{1.0}{
    {
    \setlength{\tabcolsep}{6pt}
    \begin{tabular}{@{}*2l  *8r@{}} 
    \toprule
           {Sequence} & {Public detector} & {MOTA$\uparrow$} & {IDF1$\uparrow$} & {HOTA$\uparrow$}  & {MT$\uparrow$} & {ML$\downarrow$} & {FP$\downarrow$}  & {FN$\downarrow$} & {IDs$\downarrow$}\\    \midrule  
        MOT16-01 & DPM -- CD & 65.9 & 55.6 & 45.4 & 8 & 2 & 108 & 2,040 & 34\\
        MOT16-03 & DPM -- CD & 73.6 & 67.9 & 55.8 & 78 & 15 & 3,927 & 23,475 & 166\\
        MOT16-06 & DPM -- CD & 60.7 & 61.0 & 49.2 & 92 & 29 & 1,302 & 3,083 & 154\\
        MOT16-07 & DPM -- CD & 58.9 & 51.4 & 42.7 & 14 & 8 & 481 & 6,116 & 117\\
        MOT16-08 & DPM -- CD & 46.3 & 41.2 & 35.2 & 15 & 11 & 634 & 8,182 & 174\\
        MOT16-12 & DPM -- CD & 43.9 & 58.2 & 49.0 & 27 & 12 & 1,976 & 2,616 & 64\\
        MOT16-14 & DPM -- CD & 45.3 & 45.4 & 35.6 & 25 & 37 & 1,410 & 8,4333 & 267\\
        \midrule
        \textbf{MOT16} & \textbf{DPM -- CD} & \textbf{64.5} & \textbf{60.9} & \textbf{50.3} & \textbf{259} & \textbf{114} & \textbf{9{,}838} & \textbf{53{,}945}  & \textbf{976} \\
        \bottomrule
    \end{tabular}
    }
    }
    \end{center}
    \caption{Detailed \textbf{MO3TR} results on each individual sequence of the MOT16 benchmark~\cite{mot17} test set using \textbf{public} detections. Following other works, we use the public detection filtering method using \textbf{center distances} (CD) as proposed by~\cite{tracktor}.}
    \label{tab:MOT16_detailed}
\end{table*}

\begin{table*}[h!]
    \begin{center}
    \scalebox{1.0}{
    {
    \setlength{\tabcolsep}{6pt}
    \begin{tabular}{@{}*2l  *8r@{}} 
    \toprule
           {Sequence} & {Public detector} & {MOTA$\uparrow$} & {IDF1$\uparrow$} & {HOTA$\uparrow$}  & {MT$\uparrow$} & {ML$\downarrow$} & {FP$\downarrow$}  & {FN$\downarrow$} & {IDs$\downarrow$}\\    \midrule  
        MOT16-01 & DPM - IoU & 60.7 & 51.1 & 42.8 & 7 & 4 & 88 & 2,397 & 30\\
        MOT16-03 & DPM - IoU & 73.7 & 68.0 & 55.6 & 76 & 15 & 2,817 & 24,530 & 152\\
        MOT16-06 & DPM - IoU & 54.7 & 55.8 & 45.3 & 53 & 67 & 796 & 4,311 & 119\\
        MOT16-07 & DPM - IoU & 57.2 & 50.6 & 42.0 & 14 & 8 & 409 & 6,470 & 110\\
        MOT16-08 & DPM - IoU & 44.0 & 40.2 & 34.3 & 12 & 15 & 435 & 8,784 & 155\\
        MOT16-12 & DPM - IoU & 50.5 & 60.6 & 50.3 & 26 & 17 & 1,162 & 2,901 & 44\\
        MOT16-14 & DPM - IoU & 41.3 & 43.4 & 33.7 & 20 & 44 & 1,280 & 9,298 & 270\\
        \midrule
        \textbf{MOT16} & \textbf{DPM - IoU} & \textbf{63.5} & \textbf{60.3} & \textbf{49.6} & \textbf{208} & \textbf{170} & \textbf{6{,}987} & \textbf{58{,}691}  & \textbf{880} \\
        \bottomrule
    \end{tabular}
    }
    }
    \end{center}
    \caption{Detailed \textbf{MO3TR} results on each individual sequence of the MOT16 benchmark~\cite{mot17} test set using \textbf{public} detections (DPM~\cite{felzenszwalb2009object}). Following other works, we use the public detection filtering method using \textbf{intersection over union} (IoU) as the metric as proposed in~\cite{trackformer}.}
    \label{tab:MOT16_detailed_iou}
\end{table*}

\section{Additional experimental details}
In this section, we provide detailed results for all sequences on the MOT16 and MOT17 benchmarks~\cite{mot17} as well as the MOT20~\cite{mot20} that are discussed in the main paper. For clarity, we shortly redefine the dataset characteristics and evaluation metrics that apply to these datasets.\par

\noindent \textbf{Evaluation datasets.} We use the widely established MOT16 and MOT17~\cite{mot17} datasets as well as the rather new MOT20 dataset~\cite{mot20} from the MOTchallenge benchmarks to evaluate and compare MO3TR with other state of the art models. Both MOT16 and MOT17 contain seven training and test sequences each, capturing crowded indoor or outdoor areas via moving and static cameras from various viewpoints. MOT20 contains eight sequences (four training and four test), focusing on heavily crowded scenes with a very high number of people in both day and night scenarios. Pedestrians are often heavily occluded by other pedestrians or background objects across all three datasets, making identity-preserving tracking challenging. Three sets of public detections are provided with MOT17 (DPM~\cite{felzenszwalb2009object}, FRCNN~\cite{NIPS2015_14bfa6bb} and SDP~\cite{yang2016exploit}), one with MOT16 (DPM) and one with MOT20 (FRCNN). 

\noindent \textbf{Evaluation metrics.} 
\label{sec:eval_metrics}
To evaluate MO3TR and compare its performance to other state-of-the-art tracking approaches, we use the standard set of metrics recognized by the tracking community~\cite{bernardin2008evaluating,ristani2016performance}. Analyzing the detection performance, we provide detailed insights regarding the total number of \textit{false positives}~(FP) and \textit{false negatives}~(FN, \ie missed targets). The \textit{mostly tracked targets}~(MT) measure describes the ratio of ground-truth trajectories that are covered for at least 80\% of the track's life span, while \textit{mostly lost targets}~(ML) represents the ones covered for at most 20\%. The number of \textit{identity switches} is denoted by IDs. The two most commonly used metrics to summarize the tracking performance are the \textit{multiple object tracking accuracy} (MOTA), and the identity F1 score (IDF1). MOTA combines the measures for the three error sources of false positives, false negatives and identity switches into one compact measure, and a higher MOTA score implies better performance of the respective tracking approach. The IDF1 represents the ratio of correctly identified detections over the average number of ground-truth and overall computed detections. We additionally report our results on the recently proposed \textit{higher order tracking accuracy} metric (HOTA), which aims to combine the effects of accurate detection, association and localization, and better aligns with human perception of tracking performance~\cite{luiten2021hota}. \par
All reported results are computed using the official evaluation code of the MOTChallenge benchmark.
\footnote{\url{https://motchallenge.net}}.

\subsection{MO3TR -- Evaluation results MOT16}
The public results for the MOT16~\cite{mot17} benchmark presented in the experiment section of the main paper show the overall result of MO3TR on the benchmark's test dataset using the provided public detections (DPM~\cite{felzenszwalb2009object}). Detailed results showing the individual sequence performance are presented in Table~\ref{tab:MOT16_detailed} for using the center distance to incorporate public detections as proposed in~\cite{zhou2020centertrack}, while Table~\ref{tab:MOT16_detailed_iou} states the performance achieved via the intersection over union (IoU) method~\cite{trackformer}.

\subsection{MO3TR -- Evaluation results MOT17}
The individual public results for all sequences of the MOT17 benchmark~\cite{mot17} comprising three different sets of provided public detections (DPM~\cite{felzenszwalb2009object}, FRCNN~\cite{NIPS2015_14bfa6bb} and SDP~\cite{yang2016exploit}) are detailed in Table~\ref{tab:MOT17_detailed} for using the center distance (CD) to incorporate public detections as proposed in~\cite{zhou2020centertrack}, while Table~\ref{tab:MOT17_detailed_iou} states the performance achieved via the intersection over union (IoU) method~\cite{trackformer}. The results using private detections on MOT17 are stated in Table~\ref{tab:MOT17_detailed_private_mo3tr}.

\subsection{MO3TR-PIQ -- Evaluation results MOT17 \& MOT20}
The individual results for all sequences of the MOT17 benchmark~\cite{mot17} using the private detections of our extended method MO3TR-PIQ with predicted initiation queries are detailed in Table~\ref{tab:MOT17_detailed_private_mo3tr-piq}, while the results on the MOT20 dataset~\cite{mot20} are presented in Table~\ref{tab:MOT20_detailed_private_mo3tr-piq} and Table~\ref{tab:MOT20_detailed_public_mo3tr-piq}.\par

\begin{table*}
    \begin{center}
    \scalebox{1.0}{
    {
    \setlength{\tabcolsep}{6pt}
    \begin{tabular}{@{}*2l *7r@{}} 
    \toprule
           {Sequence} & {Public detector} & {MOTA$\uparrow$} & {IDF1$\uparrow$}  & {MT$\uparrow$} & {ML$\downarrow$} & {FP$\downarrow$}  & {FN$\downarrow$} & {IDs$\downarrow$}\\    \midrule  
        MOT17-01 & DPM -- CD & 62.3 & 57.0 & 6 & 2 & 98 & 2,306 & 27\\
        MOT17-03 & DPM -- CD & 73.8 & 67.7 & 78 & 15 & 2,773 & 24,469 & 167\\
        MOT17-06 & DPM -- CD & 60.5 & 58.8 & 81 & 39 & 950 & 3,555 & 148\\
        MOT17-07 & DPM -- CD & 56.7 & 50.5 & 13 & 12 & 402 & 6,793 & 120\\
        MOT17-08 & DPM -- CD & 38.0 & 35.4 & 13 & 26 & 206 & 12,723 & 168\\
        MOT17-12 & DPM -- CD & 49.3 & 59.7 & 28 & 20 & 1,268 & 3,073 & 50\\
        MOT17-14 & DPM -- CD & 43.5 & 44.9 & 22 & 44 & 1,249 & 8,931 & 265\\
        \midrule
        MOT17-01 & FRCNN -- CD & 60.4 & 53.9 & 7 & 4 & 109 & 2,419 & 28\\
        MOT17-03 & FRCNN -- CD & 73.9 & 68.1 & 75 & 15 & 3,036 & 24,148 & 161\\
        MOT17-06 & FRCNN -- CD & 62.0 & 61.1 & 95 & 25 & 1,170 & 3,148 & 162\\
        MOT17-07 & FRCNN -- CD & 56.7 & 50.6 & 12 & 12 & 402 & 6,794 & 118\\
        MOT17-08 & FRCNN -- CD & 36.5 & 35.2 & 12 & 31 & 149 & 13,121 & 149\\
        MOT17-12 & FRCNN -- CD & 51.0 & 61.4 & 26 & 26 & 970 & 3,239 & 41\\
        MOT17-14 & FRCNN -- CD & 43.7 & 44.1 & 24 & 42 & 1,349 & 8,790 & 272\\
        \midrule
        MOT17-01 & SDP -- CD & 66.8 & 57.6 & 7 & 3 & 94 & 2,022 & 28\\
        MOT17-03 & SDP -- CD & 74.1 & 68.2 & 80 & 15 & 3,373 & 23,606 & 163\\
        MOT17-06 & SDP -- CD & 61.9 & 61.2 & 93 & 24 & 1,163 & 3,176 & 150\\
        MOT17-07 & SDP -- CD & 57.4 & 50.2 & 13 & 12 & 407 & 6,672 & 120\\
        MOT17-08 & SDP -- CD & 38.6 & 36.3 & 13 & 27 & 235 & 12,553 & 177\\
        MOT17-12 & SDP -- CD & 50.4 & 59.7 & 29 & 20 & 1,200 & 3,043 & 53\\
        MOT17-14 & SDP -- CD & 46.4 & 45.9 & 24 & 38 & 1,363 & 8,279 & 274\\
        \midrule
        \textbf{MOT17} & \textbf{All -- CD} & \textbf{63.2} & \textbf{60.2} & \textbf{751} & \textbf{452} & \textbf{21{,}966} & \textbf{182{,}860}  & \textbf{2{,}841} \\
        \bottomrule
    \end{tabular}
    }
    }
    \end{center}
    \caption{Detailed \textbf{MO3TR} results on each individual sequence of the MOT17 benchmark~\cite{mot17} test set using \textbf{public} detections (DPM~\cite{felzenszwalb2009object}, FRCNN~\cite{NIPS2015_14bfa6bb}, SDP~\cite{yang2016exploit}). Following other works, we use the public detection filtering method using \textbf{center distances} (CD) as proposed by~\cite{tracktor}.}
    \label{tab:MOT17_detailed}
\end{table*}

\begin{table*}
    \begin{center}
    \scalebox{1.0}{
    {
    \setlength{\tabcolsep}{6pt}
    \begin{tabular}{@{}*2l *8r@{}} 
    \toprule
           {Sequence} & {Public detector} & {MOTA$\uparrow$} & {IDF1$\uparrow$} & {HOTA$\uparrow$}  & {MT$\uparrow$} & {ML$\downarrow$} & {FP$\downarrow$}  & {FN$\downarrow$} & {IDs$\downarrow$}\\    \midrule  
        MOT17-01 & DPM -- IoU & 60.2 & 50.9 & 42.0 & 6 & 5 & 87 & 2,452 & 30\\
        MOT17-03 & DPM -- IoU & 73.7 & 68.0 & 55.8 & 76 & 15 & 2,779 & 24,603 & 152\\
        MOT17-06 & DPM -- IoU & 55.5 & 55.8 & 45.3 & 54 & 66 & 680 & 4,441 & 120\\
        MOT17-07 & DPM -- IoU & 55.6 & 49.8 & 41.4 & 13 & 12 & 373 & 7,007 & 113\\
        MOT17-08 & DPM -- IoU & 37.2 & 35.4 & 30.8 & 12 & 29 & 174 & 12,924 & 163\\
        MOT17-12 & DPM -- IoU & 50.8 & 60.5 & 50.0 & 28 & 20 & 1,048 & 3,168 & 45\\
        MOT17-14 & DPM -- IoU & 41.3 & 43.4 & 33.7 & 20 & 44  & 1,280 & 9,298 & 270\\
        \midrule
        MOT17-01 & FRCNN -- IoU &  60.9 & 53.5 & 43.6 & 7 & 3 & 102 &  2,391 & 28\\
        MOT17-03 & FRCNN -- IoU & 73.6 & 68.1 & 55.8 & 77 & 15 & 3,973 & 24,458 & 160\\
        MOT17-06 & FRCNN -- IoU & 58.8 & 57.2 & 46.3 & 70 & 48 & 920 & 3,793 & 137\\
        MOT17-07 & FRCNN -- IoU & 55.8 & 50.6 & 41.7 & 11 & 12 & 368 & 6,991 & 111\\
        MOT17-08 & FRCNN -- IoU & 36.4 & 35.3 & 30.8 & 12 & 31 & 150 & 13,138 & 144\\
        MOT17-12 & FRCNN -- IoU & 51.8 & 61.7 & 50.2 & 26 & 28 & 813 & 3,329 & 36\\
        MOT17-14 & FRCNN -- IoU & 43.0 & 43.9 & 33.8 & 21 & 40  & 1,247 & 9,021 & 268\\
        \midrule
        MOT17-01 & SDP -- IoU & 66.6 & 57.6 & 46.1 & 7 & 3 & 94 & 2,031 & 28\\
        MOT17-03 & SDP -- IoU & 74.1 & 68.1 & 56.1 & 80 & 15 & 3,348 & 23,612 & 164\\
        MOT17-06 & SDP -- IoU & 58.6 & 57.7 & 46.8 & 70 & 53 & 868 & 3,888 & 126\\
        MOT17-07 & SDP -- IoU & 56.6 & 50.0 & 41.8 & 13 & 14 & 380 & 6,835 & 113\\
        MOT17-08 & SDP -- IoU & 38.4 & 35.8 & 31.2 & 13 & 28 & 216 & 12,626 & 176\\
        MOT17-12 & SDP -- IoU & 52.2 & 61.3 & 50.4 & 27 & 22 & 906 & 3,196 & 40\\
        MOT17-14 & SDP -- IoU & 46.4 & 45.8 & 36.0 & 26 & 44  & 1,269 & 8,376 & 256\\
        \midrule
        \textbf{MOT17} & \textbf{All -- IoU} & \textbf{62.70} & \textbf{59.90} & \textbf{49.40} & \textbf{669} & \textbf{547} & \textbf{20{,}075} & \textbf{187{,}578}  & \textbf{2{,}680} \\
        \bottomrule
    \end{tabular}
    }
    }
    \end{center}
    \caption{Detailed \textbf{MO3TR} results on each individual sequence of the MOT17 benchmark~\cite{mot17} test set using \textbf{public} detections (DPM~\cite{felzenszwalb2009object}, FRCNN~\cite{NIPS2015_14bfa6bb}, SDP~\cite{yang2016exploit}). Following other works, we use the public detection filtering method using \textbf{intersection over union} (IoU) as proposed in~\cite{trackformer}.}
    \label{tab:MOT17_detailed_iou}
\end{table*}

\begin{table*}
    \begin{center}
    \scalebox{1.0}{
    {
    \setlength{\tabcolsep}{6pt}
    \begin{tabular}{@{}*2l *8r@{}} 
    \toprule
           {Sequence} & {MOTA$\uparrow$} & {IDF1$\uparrow$} & {HOTA$\uparrow$}  & {MT$\uparrow$} & {ML$\downarrow$} & {FP$\downarrow$}  & {FN$\downarrow$} & {IDs$\downarrow$}\\    \midrule  
        MOT17-01 & 66.9 & 57.9 & 46.3 & 7 & 3 & 94 & 2,012 & 28\\
        MOT17-03 & 74.0 & 68.1 & 56.1 & 80 & 15 & 3,488 & 23,573 & 165\\
        MOT17-06 & 61.1 & 60.8 & 49.3 & 93 & 25 & 1,080 & 3,361 & 143\\
        MOT17-07 & 57.4 & 50.6 & 42.1 & 13 & 13 & 385 & 6,688 & 116\\
        MOT17-08 & 39.6 & 36.6 & 31.5 & 13 & 25 & 256 & 12,327 & 186\\
        MOT17-12 & 52.2 & 61.0 & 50.4 & 29 & 17 & 1,094 & 3,000 & 51\\
        MOT17-14 & 46.3 & 45.8 & 36.2 & 28 & 42  & 1,389 & 8,267 & 269\\
        \midrule
        \textbf{MOT17} & \textbf{63.9} & \textbf{60.5} & \textbf{49.9} & \textbf{789} & \textbf{420} & \textbf{23{,}358} & \textbf{177{,}684}  & \textbf{2{,}874} \\
        \bottomrule
    \end{tabular}
    }
    }
    \end{center}
    \caption{Detailed \textbf{MO3TR} results on each individual sequence of the MOT17 benchmark~\cite{mot17} test set using our own \textbf{private} detections and our MO3TR with learnt initiation queries.}
    \label{tab:MOT17_detailed_private_mo3tr}
\end{table*}

\begin{table*}
    \begin{center}
    \scalebox{1.0}{
    {
    \setlength{\tabcolsep}{6pt}
    \begin{tabular}{@{}*2l *8r@{}} 
    \toprule
           {Sequence} & {MOTA$\uparrow$} & {IDF1$\uparrow$} & {HOTA$\uparrow$}  & {MT$\uparrow$} & {ML$\downarrow$} & {FP$\downarrow$}  & {FN$\downarrow$} & {IDs$\downarrow$}\\    \midrule  
        \revised{MOT17-01} & \revised{60.4} & \revised{47.6} & \revised{44.1} & \revised{13} & \revised{5} & \revised{402} & \revised{2,109} & \revised{41}\\
        \revised{MOT17-03} & \revised{91.7} & \revised{85.4} & \revised{69.7} & \revised{141} & \revised{0} & \revised{3,147} & \revised{5,370} & \revised{139}\\
        \revised{MOT17-06} & \revised{60.8} & \revised{59.0} & \revised{48.3} & \revised{75} & \revised{52} & \revised{509} & \revised{3,982} & \revised{129}\\
        \revised{MOT17-07} & \revised{69.4} & \revised{53.8} & \revised{46.4} & \revised{27} & \revised{4} & \revised{529} & \revised{4,504} & \revised{134}\\
        \revised{MOT17-08} & \revised{57.8} & \revised{46.0} & \revised{41.4} & \revised{25} & \revised{8} & \revised{821} & \revised{7,837} & \revised{260}\\
        \revised{MOT17-12} & \revised{60.2} & \revised{67.8} & \revised{56.4} & \revised{37} & \revised{12} & \revised{1,151} & \revised{2,250} & \revised{49}\\
        \revised{MOT17-14} & \revised{52.5} & \revised{58.3} & \revised{43.0} & \revised{21} & \revised{32} & \revised{456} & \revised{8,125} & \revised{197}\\
        \midrule
        \revised{\textbf{MOT17}} & \revised{\textbf{77.6}} & \revised{\textbf{72.9}} & \revised{\textbf{60.3}} & \revised{\textbf{1,017}} & \revised{\textbf{339}} & \revised{\textbf{21{,}045}} & \revised{\textbf{102{,}531}} & \revised{\textbf{2{,}847}} \\
        \bottomrule
    \end{tabular}
    }
    }
    \end{center}
    \caption{\revised{Detailed \textbf{MO3TR-PIQ} results on each individual sequence of the MOT17 benchmark~\cite{mot17} test set using our own \textbf{private} detections and initiation queries predicted via our modified MO3TR version.}}
    \label{tab:MOT17_detailed_private_mo3tr-piq}
\end{table*}

\begin{table*}
    \begin{center}
    \scalebox{1.0}{
    {
    \setlength{\tabcolsep}{6pt}
    \begin{tabular}{@{}*2l *8r@{}} 
    \toprule
           {Sequence} & {MOTA$\uparrow$} & {IDF1$\uparrow$} & {HOTA$\uparrow$}  & {MT$\uparrow$} & {ML$\downarrow$} & {FP$\downarrow$}  & {FN$\downarrow$} & {IDs$\downarrow$}\\    \midrule  
        MOT20-04 & 85.0 & 78.4 & 65.0 & 491 & 21 & 7,351 & 32,749 & 1,063\\
        MOT20-06 & 58.1 & 55.7 & 46.1 & 96 & 83 & 2,644	 & 52,394 & 583\\
        MOT20-07 & 79.9 & 67.6 & 60.1 & 82 & 3 & 1,245 & 5,201 & 221\\
        MOT20-08 & 48.5 & 52.3 & 41.3 & 37 & 73 & 1,498 & 38,095 & 333\\
        \midrule
        \textbf{MOT20} & \textbf{72.3} & \textbf{69.0} & \textbf{57.3} & \textbf{706} & \textbf{180} & \textbf{12,738} & \textbf{128,439}  & \textbf{2,200} \\
        \bottomrule
    \end{tabular}
    }
    }
    \end{center}
    \caption{Detailed \textbf{MO3TR-PIQ} results on each individual sequence of the MOT20 benchmark~\cite{mot20} test set using our own \textbf{private} detections and initiation queries predicted via our modified MO3TR version.}
    \label{tab:MOT20_detailed_private_mo3tr-piq}
\end{table*}

\begin{table*}
    \begin{center}
    \scalebox{1.0}{
    {
    \setlength{\tabcolsep}{6pt}
    \begin{tabular}{@{}*2l *8r@{}} 
    \toprule
           {Sequence} & {MOTA$\uparrow$} & {IDF1$\uparrow$} & {HOTA$\uparrow$}  & {MT$\uparrow$} & {ML$\downarrow$} & {FP$\downarrow$}  & {FN$\downarrow$} & {IDs$\downarrow$}\\    \midrule  
        \revised{MOT20-04} & \revised{83.6} & \revised{78.5} & \revised{64.3} & \revised{459} & \revised{41} & \revised{4,649} & \revised{39,821} & \revised{578}\\
        \revised{MOT20-06} & \revised{50.8} & \revised{51.6} & \revised{40.9} & \revised{60} & \revised{100} & \revised{777}	 & \revised{64,105} & \revised{972}\\
        \revised{MOT20-07} & \revised{72.5} & \revised{65.8} & \revised{55.9} & \revised{53} & \revised{9} & \revised{779} & \revised{8,157} & \revised{217}\\
        \revised{MOT20-08} & \revised{37.0} & \revised{40.7} & \revised{34.0} & \revised{20} & \revised{94 }& \revised{348} & \revised{48,340} & \revised{386}\\
        \midrule
        \revised{\textbf{MOT20} } & \revised{ \textbf{67.5} } & \revised{\textbf{66.9} } & \revised{ \textbf{54.9} } & \revised{ \textbf{592} } & \revised{ \textbf{244} } & \revised{ \textbf{6,553} } & \revised{ \textbf{160,423} } & \revised{ \textbf{1,280}} \\
        \bottomrule
    \end{tabular}
    }
    }
    \end{center}
    \caption{\revised{Detailed \textbf{MO3TR-PIQ} results on each individual sequence of the MOT20 benchmark~\cite{mot20} test set using the provided \textbf{public} detections. Following other works, we use the public detection filtering method using \textbf{intersection over union} (IoU) as proposed in~\cite{trackformer}.}}
    \label{tab:MOT20_detailed_public_mo3tr-piq}
\end{table*}




\end{document}